\pdfoutput=1

\documentclass[10pt,twocolumn,letterpaper]{article}

\usepackage[pagenumbers]{cvpr}   %

\newcommand{\suppsec}[1]{Appendix~#1}
\newcommand{\ofmain}{}

\def\ppmode{}
\newcommand{\mref}[2]{\ref{#1}}

\usepackage{booktabs}
\usepackage{multirow}
\usepackage{colortbl}
\usepackage{xcolor}
\usepackage{amsmath,amssymb}
\usepackage{graphicx}
\usepackage{xspace}
\usepackage{pifont}
\usepackage{tikz}
\usetikzlibrary{arrows.meta,calc}
\providecommand{\suppsec}[1]{Section~#1 of the supplementary material}

\providecommand{\mref}[2]{#2}   %
\providecommand{\ofmain}{ of the main paper}
\newcommand{\ours}{M\textsuperscript{3}OPD-Reasoner\xspace}
\newcommand{\oursmodel}{M\textsuperscript{3}OPD-Reasoner-9B\xspace}
\newcommand{\oursft}{M\textsuperscript{3}OPD-SFT-1.35M\xspace}
\newcommand{\ourrl}{M\textsuperscript{3}OPD-RL-193K\xspace}
\newcommand{\oursfull}{Multimodal Model Multi-Teacher On-Policy Distillation Reasoner\xspace}
\IfFileExists{figures/colors.tex}{\definecolor{domImgr}{HTML}{3775BA}   %
\definecolor{domImgp}{HTML}{DE8A2A}   %
\definecolor{domVid}{HTML}{2A9A78}    %
\definecolor{domImgrL}{HTML}{DDE8F5}
\definecolor{domImgpL}{HTML}{FBE7D1}
\definecolor{domVidL}{HTML}{D4EEE5}
\definecolor{figInk}{HTML}{272727}
\definecolor{figGray}{HTML}{5C5C5C}
\definecolor{figLine}{HTML}{CFCECE}
\definecolor{figPanel}{HTML}{F6F6F4}
\definecolor{figStudent}{HTML}{4D4D4D}
\definecolor{figStudentL}{HTML}{E9E9E9}
\definecolor{figGold}{HTML}{F2C14E}
}{}
\IfFileExists{tables/macros.tex}{\input{tables/macros.tex}}{}
\IfFileExists{tables/results_macros.tex}{

}{}
\IfFileExists{tables/mixrl_macros.tex}{

\providecommand{\mxFullN}{193,027}

}{}
\IfFileExists{tables/compare_macros.tex}{

}{}

\providecommand{\todo}[1]{{\color{red}#1}}

\providecolor{oursL}{HTML}{EEF3FA}   %
\providecolor{overallL}{HTML}{E8E8E8}
\providecommand{\svAccW}{0.9}
\providecommand{\svFmtW}{0.1}
\providecommand{\svPartialBox}{0.5}
\providecommand{\svAnls}{0.5}
\providecommand{\svCoordScale}{1000}
\providecommand{\svPointRadius}{20}
\providecommand{\svMaxAnsChars}{300}
\providecommand{\svMathTimeout}{3}
\providecommand{\svNgram}{13}
\providecommand{\svPhash}{4}
\providecommand{\svEvalCap}{32768}
\providecommand{\svEvalT}{0.6}
\providecommand{\svEvalTopP}{0.95}
\providecommand{\svEvalN}{8}
\providecommand{\svEvalTopK}{20}
\providecommand{\svImgMin}{65,536}
\providecommand{\svImgMax}{1,605,632}
\providecommand{\svFps}{1}
\providecommand{\svFrMin}{4}
\providecommand{\svFrMax}{150}
\providecommand{\svFrPxMin}{4,096}
\providecommand{\svFrPxMax}{65,536}
\providecommand{\svTopK}{32}
\providecommand{\svClamp}{10}
\providecommand{\svLpMin}{$-10$}
\providecommand{\svMopdBatch}{1008}
\providecommand{\svMopdSteps}{191}
\providecommand{\svMopdPrompts}{193,027}
\providecommand{\svMopdResp}{16384}
\providecommand{\svStale}{2}
\providecommand{\svTeacherMml}{22017}
\providecommand{\svMopdPrompt}{5632}
\providecommand{\svGspoN}{8}
\providecommand{\svSftTokStep}{1,048,576}
\providecommand{\svVerlVer}{0.9.1}
\providecommand{\svVllmVer}{0.19.1}
\providecommand{\svEngineVer}{0.1.4}
\providecommand{\svEvalVer}{0.7.2}
\providecommand{\svMvVer}{0.9.0}
\providecommand{\svTorchVer}{2.10.0}
\providecommand{\svPoolSeed}{20261022}
\providecommand{\svGenSelfT}{0.7}
\providecommand{\svKeepSelf}{78.1}
\providecommand{\svKeepNext}{99.1}
\providecommand{\svMixRLStep}{240}
\makeatletter\let\pre@paragraph\paragraph
\renewcommand\paragraph{\@startsection{paragraph}{4}{\z@}{0.8ex \@plus0.4ex \@minus0.2ex}{-1em}{\normalfont\normalsize\bfseries}}\makeatother
\usepackage{algorithm}
\usepackage{algpseudocode}
\usepackage{float}

\definecolor{cvprblue}{rgb}{0.21,0.49,0.74}
\usepackage[pagebackref,breaklinks,colorlinks,allcolors=cvprblue]{hyperref}
\hypersetup{pdftitle={M3OPD-Reasoner: An Open Recipe for Post-Training Multimodal Reasoning with Multi-Teacher On-Policy Distillation},
  pdfauthor={Nguyen Quang Trung, Hoang Phan, Shuo Sun, Longyin Zhang, AiTi Aw}}

\def\confName{CVPR}
\def\confYear{2027}

\newcommand{\urlpage}{https://ngquangtrung57.github.io/m3opd-reasoner-page/}
\newcommand{\urlcode}{https://github.com/ngquangtrung57/M3OPD-Reasoner}
\newcommand{\urlsft}{https://huggingface.co/datasets/ngqtrung/M3OPD-SFT-1.35M}
\newcommand{\urlrl}{https://huggingface.co/datasets/ngqtrung/M3OPD-RL-193K}
\newcommand{\urlcoll}{https://huggingface.co/collections/ngqtrung/m3opd-reasoner-6aed5164d151b6643a1b3b2d}

\title{\ours: An Open Recipe for Post-Training Multimodal Reasoning\\with Multi-Teacher On-Policy Distillation}

\author{%
  Nguyen Quang Trung\textsuperscript{1}\quad Hoang Phan\textsuperscript{2}\quad Shuo Sun\textsuperscript{1}\quad Longyin Zhang\textsuperscript{1}\quad AiTi Aw\textsuperscript{1}\\[0.2em]
  \textsuperscript{1}A*STAR\qquad \textsuperscript{2}VinUniversity\\[0.4em]
  {\normalsize
  \href{\urlpage}{Project Page}\quad$\cdot$\quad
  \href{\urlcode}{Code}\quad$\cdot$\quad
  Data: \href{\urlsft}{SFT}, \href{\urlrl}{RL}\quad$\cdot$\quad
  \href{\urlcoll}{Collection}}%
}

\makeatletter
\newcommand{\suppstart}{%
  \onecolumn
  \let\paragraph\pre@paragraph
  \raggedbottom
  \let\supp@xfloat\@xfloat
  \def\@xfloat##1[##2]{\supp@xfloat{##1}[!htbp]}%
  \renewenvironment{table*}[1][]{\table[!htbp]}{\endtable}%
  \renewenvironment{figure*}[1][]{\figure[!htbp]}{\endfigure}%
  \renewcommand{\topfraction}{0.9}\renewcommand{\bottomfraction}{0.8}\renewcommand{\textfraction}{0.07}\renewcommand{\floatpagefraction}{0.85}%
  \setcounter{topnumber}{3}\setcounter{bottomnumber}{2}\setcounter{totalnumber}{5}%
  \renewcommand*\l@section{\@dottedtocline{1}{0pt}{1.72em}}%
  \def\theHtable{S\arabic{table}}%
  \def\theHfigure{S\arabic{figure}}%
  \def\theHequation{S\arabic{equation}}%
  \def\theHalgorithm{S\arabic{algorithm}}%
  \addtocontents{toc}{\protect\setcounter{tocdepth}{1}}%
  \begin{center}\Large\textbf{Appendix}\end{center}%
}
\makeatother
\AtBeginDocument{\addtocontents{toc}{\protect\setcounter{tocdepth}{-10}}}

\begin{document}
\twocolumn[{\renewcommand\twocolumn[1][]{#1}\maketitle
\begin{center}
  \centering
  \captionsetup{type=figure}
  \includegraphics[width=0.96\textwidth]{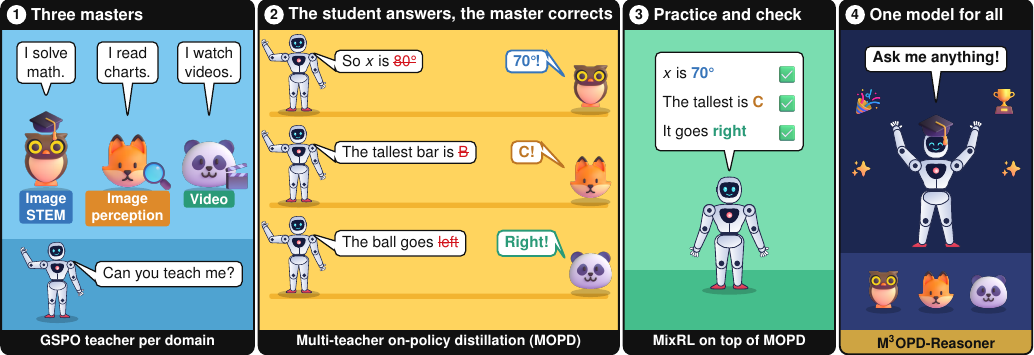}
  \captionof{figure}{\oursfull (\ours) as a story. (1) Three masters train with GSPO, one per domain, with or without an SFT warm-up. (2) The student answers on its own, and the master of the prompt's domain corrects each token (MOPD). (3) The student practices on all domains, and a rule checks each answer (MixRL). (4) One model serves all three domains.}
  \label{fig:teaser}
\end{center}
\vspace{0.5em}
}]
\begin{abstract}
Multi-Teacher On-Policy Distillation (MOPD) enables a single student to absorb the skills of several domain teachers by learning from its own generated responses.
In this work, we present \oursfull (\ours), a fully open and complete recipe for post-training vision-language models (VLMs) over three reasoning domains: image STEM, image perception, and video.
We first curate two datasets across the three domains, \oursft for supervised fine-tuning (SFT) and \ourrl for reinforcement learning (RL) with verifiable rewards.
On these data, we compare post-training recipes that start from Qwen3.5-9B: training one model on the mixed data of all domains (MixSFT and MixRL), and training one teacher per domain with Group Sequence Policy Optimization (GSPO), directly or after an SFT warm-up, and distilling the teachers into one student with MOPD.
We evaluate on 21 image STEM, image perception, and video reasoning benchmarks after we decontaminate the training data against them.
The comparison establishes MOPD of teachers trained with GSPO directly from Qwen3.5-9B as the best recipe: the student closely matches each teacher on its own domain and is on average above the mixed-training baselines on the same prompts.
Since MOPD already merges the domain capabilities that MixRL would otherwise have to learn, its student is a stronger starting point for RL, and we ask whether the same data can carry the model further from there: we add MixRL on top of the MOPD student, on the same prompts and with the same settings, which, to our knowledge, no prior VLM recipe has studied.
The same prompts still improve the MOPD student, with no collapse, and give our main model, \oursmodel.
MOPD and MixRL can therefore work as consecutive stages that get more out of the same data.
To support reproducible research, we \ifdefined\ppmode release\else will release\fi{} all data, the training code, and the checkpoints of every stage.
\end{abstract}

\ifdefined\ppmode\else\begin{figure*}[t]
  \centering
  \IfFileExists{figures/pipeline.pdf}{%
    \includegraphics[width=\textwidth]{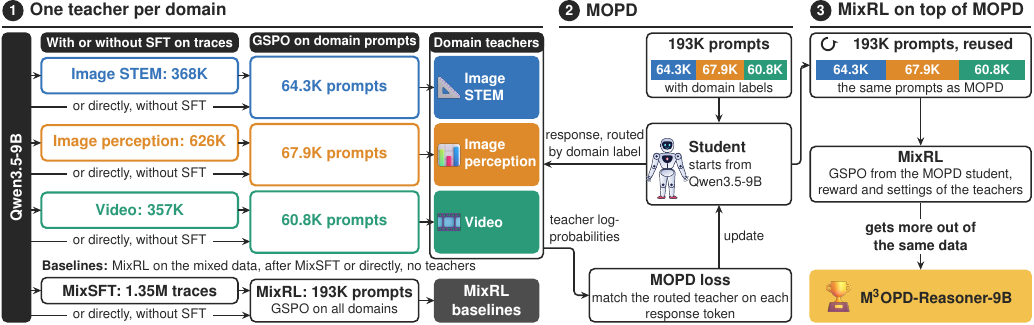}%
  }{%
    \fbox{\parbox[c][4cm][c]{0.95\textwidth}{\centering\todo{figures/pipeline.pdf missing}}}%
  }
  \caption{Overview of the recipe. (1) GSPO on the prompts of one domain turns Qwen3.5-9B into the teacher of that domain, directly or after an SFT warm-up on the traces of the domain. (2) MOPD distills the domain teachers into one student that starts from Qwen3.5-9B, on the union of the prompts. (3) MixRL continues the MOPD student on the same prompts and gives the main model, \oursmodel. The baselines train one model on the mixed data with MixRL, after MixSFT or directly from Qwen3.5-9B, without teachers or MOPD.}
  \label{fig:pipeline}
\end{figure*}
\fi

\section{Introduction}\label{sec:intro}

Open recipes for reasoning in vision-language models (VLMs) have advanced quickly: OpenMMReasoner~\cite{zhang2025openmmreasoner}, Video-R1~\cite{feng2025videor1}, Video-Thinker~\cite{wang2025videothinker}, OneThinker~\cite{feng2025onethinker} and Vero~\cite{sarch2026vero} all report strong results on image or video benchmarks.
Most of them follow two phases, which have been common since DeepSeek-R1~\cite{guo2025deepseekr1}.
A supervised fine-tuning (SFT) phase, often on a mix of domains (MixSFT), gives the model a cold start and teaches it a reasoning style, and reinforcement learning (RL) with verifiable rewards on mixed-domain prompts (MixRL) then strengthens its reasoning.
Vero shows that MixRL alone, on a large mixed set of 600K samples, also works without SFT, and recent VLM releases similarly train on the union of the domain data~\cite{hong2025glm41v,bai2025qwen3vl}.
Another route trains one RL expert per domain and then merges their weights~\cite{ilharco2023taskarithmetic,yadav2023ties}, and recent text studies often compare this route with MixRL~\cite{wang2026mixmerge,wu2026consolidating}.

More recently, On-Policy Distillation (OPD) has become a better alternative to off-policy distillation~\cite{agarwal2024gkd,gu2024minillm}.
In OPD, the student samples its own responses and the teacher scores every token of the response, which reduces the exposure bias of training on teacher-generated text.
Multi-Teacher On-Policy Distillation (MOPD) extends OPD to one RL teacher per domain, where each prompt goes to the teacher of its domain, and has become a common post-training method to unify the capabilities of one model~\cite{xiaomi2026mimov2flash,ma2026mopd,yang2026nemotroncascade2}.
Building on this, several studies report that MOPD outperforms MixRL on text~\cite{ma2026mopd} and on images~\cite{lin2026mmvistareason,xiong2026scoped}, and also weight merging~\cite{ma2026mopd,xiong2026scoped}.
They give two reasons: in MixRL, a gain in one domain can come with a loss in another~\cite{ma2026mopd}, and the gradients of different capabilities conflict~\cite{xiong2026scoped}, whereas MOPD keeps the skill of each domain in a dedicated teacher.

However, this recipe is not yet well established for reasoning VLMs.
MMVistaReason~\cite{lin2026mmvistareason} also goes from SFT through GSPO experts to MOPD, but it covers images only.
SCOPED~\cite{xiong2026scoped} also distills image-skill experts only, and Video-MOPD~\cite{qin2026videomopd} adds video teachers but compares only against model merging.
These recipes end at distillation, and other multimodal students add the reward inside the distillation stage~\cite{zhao2026opod,lian2026uimopd}.

Moreover, a comparison of MOPD and MixRL as competing routes misses how the two relate.
In the two-phase recipe, MixRL starts from the base or SFT model and has two tasks at once: it must merge the capabilities of all domains into one policy, and it must push the reasoning of that policy further.
Since MOPD has proven more effective at the first of these tasks~\cite{ma2026mopd,lin2026mmvistareason}, its student offers a stronger starting point for RL than the base or SFT model.
This raises a question that the comparison leaves open: from this new starting point, can RL on the same data still carry the model further?
To our knowledge, such a separate reward-driven stage after MOPD remains unstudied for VLMs.

In this work, we present \textbf{\oursfull (\ours)}, a fully open recipe for reasoning VLMs across image STEM reasoning (math, science, and logic problems), image perception reasoning (charts, documents, counting, and spatial questions), and video reasoning (Figure~\ref{fig:pipeline}).
We first build the data of each domain (Section~\ref{sec:recipe_data}), and on these data, starting from Qwen3.5-9B~\cite{qwen2026qwen35}, we compare post-training recipes.
The mixed-training recipes train one model on the union of the domain data with MixSFT, MixRL, or both.
The multi-teacher recipe trains one teacher per domain with Group Sequence Policy Optimization (GSPO)~\cite{zheng2025gspo} on the RL prompts of that domain, directly or after an SFT warm-up, and then distills the three teachers into one student with MOPD.
The comparison establishes MOPD of teachers trained with GSPO directly from Qwen3.5-9B as the best recipe (Section~\ref{sec:main_results}).
Finally, we initialize MixRL from the MOPD student and train it on the same prompts with the same GSPO settings. This additional stage further improves the student and yields our main model, \oursmodel.
Beyond the final model, we \ifdefined\ppmode release\else will release\fi{} every part of the recipe, from the data and the training code to the checkpoints of all stages.

Our contributions are as follows:
\begin{itemize}
  \item We build a fully open recipe for the three reasoning domains \ifdefined\ppmode and release\else and will release\fi{} all of it: the data, with \oursft (1.35M samples for SFT) and \ourrl (193K prompts with verifiable answers for RL), the training code, and the checkpoints of every stage.
  \item We compare post-training recipes for multimodal reasoning on the same data, from mixed training with MixSFT and MixRL to MOPD of GSPO domain teachers with or without an SFT warm-up, and we establish MOPD of teachers trained directly from Qwen3.5-9B as the best recipe: its student matches each teacher on its own domain and is above the mixed-training baselines overall.
  \item We study MixRL on top of the MOPD student as a final stage, which starts RL from the stronger point that MOPD provides instead of from the base model. To our knowledge, this is the first study of a separate RL stage after multi-teacher distillation in VLMs. On the same prompts and settings, the stage still improves the MOPD student with no collapse, and we analyze how the stronger start changes the course of MixRL.
\end{itemize}

\section{Related work}\label{sec:related}
\ifdefined\ppmode\begin{figure*}[t]
  \centering
  \IfFileExists{figures/pipeline.pdf}{%
    \includegraphics[width=\textwidth]{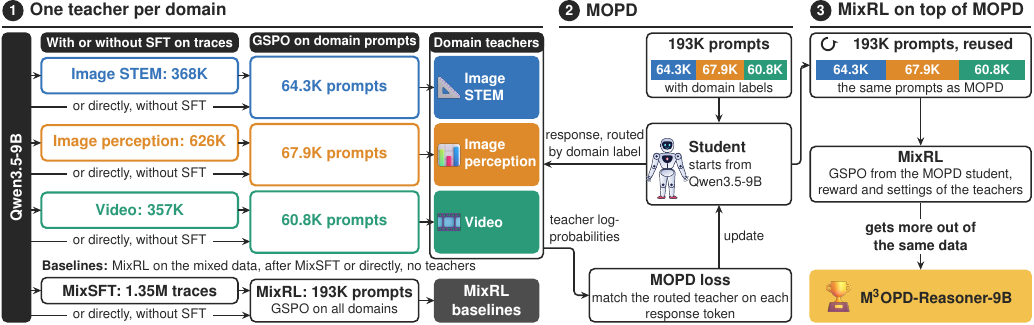}%
  }{%
    \fbox{\parbox[c][4cm][c]{0.95\textwidth}{\centering\todo{figures/pipeline.pdf missing}}}%
  }
  \caption{Overview of the recipe. (1) GSPO on the prompts of one domain turns Qwen3.5-9B into the teacher of that domain, directly or after an SFT warm-up on the traces of the domain. (2) MOPD distills the domain teachers into one student that starts from Qwen3.5-9B, on the union of the prompts. (3) MixRL continues the MOPD student on the same prompts and gives the main model, \oursmodel. The baselines train one model on the mixed data with MixRL, after MixSFT or directly from Qwen3.5-9B, without teachers or MOPD.}
  \label{fig:pipeline}
\end{figure*}
\fi%

\begin{figure*}[!t]
  \centering
  {\input{figures/data_mix}}
  \caption{Data sources of (a) \oursft and (b) \ourrl.}
  \label{fig:data}
\end{figure*}

\paragraph{On-policy and multi-teacher distillation.}
On-policy distillation~\cite{agarwal2024gkd,gu2024minillm} is now a standard post-training stage for small models~\cite{yang2025qwen3}.
For VLMs, single-teacher variants reweight visually grounded tokens~\cite{liu2026vaopd}, steer the update toward visual grounding~\cite{yoon2026decomposedopd}, \ifdefined\ppmode transfer reasoning from a text teacher~\cite{bousselham2025vold}, or use the same model with a crop of the evidence region as the teacher~\cite{yuan2026visionopd}\else or transfer reasoning from a text teacher~\cite{bousselham2025vold}\fi.
MOPD extends this setting to several domain teachers~\cite{xiaomi2026mimov2flash,ma2026mopd,yang2026nemotroncascade2}.
Later studies address the imbalance between domains~\cite{gao2026openmopd,sun2026d3mopd}, arbitrate between teachers per token~\cite{yin2026hopd}, or protect general skills~\cite{yin2026sfmopd}, and one compares it with off-policy SFT on teacher outputs~\cite{xie2026rethinking}.
Multi-teacher recipes for VLMs distill image-skill experts~\cite{xiong2026scoped}, video and image experts~\cite{qin2026videomopd}, and RL teachers for text, image, video and agent tasks~\cite{keye2026keyevl2}.
The closest work is MMVistaReason~\cite{lin2026mmvistareason}, which trains image-reasoning GSPO experts on Qwen3.5 and distills them into one student by MOPD, with open data.
In comparison, we also cover video, and we study a MixRL stage after MOPD.

\paragraph{Open RL recipes for reasoning.}
Reinforcement learning with verifiable rewards drives recent gains in LLM reasoning, with group-based methods such as Group Relative Policy Optimization (GRPO)~\cite{shao2024deepseekmath,guo2025deepseekr1} and the sequence-level GSPO~\cite{zheng2025gspo}.
Open VLM recipes apply these methods to image reasoning~\cite{meng2025mmeureka,wang2025vlrethinker,leng2025mmr1,zhang2025openmmreasoner,sarch2026vero}, to video understanding~\cite{feng2025videor1,wang2025videorft,li2025videochatr1}, and to both in one model~\cite{feng2025onethinker}.
We use per-domain GSPO only to build the teachers, and we use MixRL both as a baseline and as a new final stage on top of the distilled student.
A few text recipes also continue with RL after MOPD: Nemotron-Cascade~2 runs further domain RL stages~\cite{yang2026nemotroncascade2}, and iterated MOPD retrains per-domain teachers from the student~\cite{ma2026mopd}.
For single-teacher text OPD, the distilled model is also a stronger start for later RL than the SFT model~\cite{acikgoz2026synergy}.
For omni-modal and GUI students, OPOD and UI-MOPD instead optimize a reward inside the distillation stage~\cite{zhao2026opod,lian2026uimopd}, and the multi-teacher VLM recipes above end at MOPD.
To our knowledge, our work is the first to study a separate RL stage after multi-teacher distillation in VLMs.

\section{Recipe}\label{sec:recipe}

We present our recipe, which consolidates three domain teachers into one vision-language model by Multi-Teacher On-Policy Distillation (MOPD), over three reasoning domains: image STEM, image perception, and video.
All models build on Qwen3.5-9B~\cite{qwen2026qwen35}, and, as shown in Figure~\ref{fig:pipeline}, the recipe has three stages: (1) domain teachers, each trained with GSPO, (2) MOPD of the three teachers into one student, and (3) MixRL on top of the MOPD student.

\subsection{Data}\label{sec:recipe_data}

\paragraph{Domains and sources.}
Each domain has an SFT set and an RL prompt set, built from public datasets.
We show every source with its size in Figure~\ref{fig:data} and provide more details in \suppsec{A}.
Image STEM covers math, science, and visual logic problems, mainly from MMFineReason~\cite{lin2026mmfinereason}, MMVistaReason~\cite{lin2026mmvistareason}, DeepVision-103K~\cite{sun2026deepvision}, the STEM part of Vero~\cite{sarch2026vero}, and ViRL39K~\cite{wang2025vlrethinker}.
Image perception covers charts, documents, OCR, counting, grounding, and spatial questions, mainly from FineVision~\cite{wiedmann2025finevision}, ChartVerse~\cite{liu2026chartverse}, ChartNet~\cite{kondic2026chartnet}, and the perception parts of Vero.
Finally, video comes mainly from VideoChat3~\cite{li2026videochat3}, LLaVA-Video-178K~\cite{zhang2024llavavideo}, OneThinker~\cite{feng2025onethinker}, LongVideo-Reason~\cite{chen2025longvideo}, Video-HopChain~\cite{quang2026videohopchain}, CapRL-Video-QA~\cite{yang2026caprlpp}, and TimeLens~\cite{zhang2026timelens,zhu2026timelens2}.

\paragraph{SFT traces.}
The SFT sets serve the MixSFT baseline and the SFT warm-up of the teachers that we compare (Section~\ref{sec:recipe_teachers}).
We replace each response of the SFT sources with a new sample from a model that we give the gold final answer.
This model, the generator, receives the media, the question, and the gold answer, and it writes a thinking trace and a reply that ends in that answer.
After generation, we drop a trace if its final answer differs from the gold answer, if it refers to the given answer, if the generator rejects the gold, or if it is truncated or degenerate.

\ifdefined\ppmode\begin{table}[t]
\centering\footnotesize
\setlength{\tabcolsep}{4pt}
\renewcommand{\arraystretch}{1.05}
\caption{SFT on Qwen3.8-Flash-Next traces and on self traces of Qwen3.5-9B (domain mean of the SFT model of each domain on the benchmarks of its domain, avg@8 accuracy, \%). $\Delta$ is the change from the left to the right value.}
\label{tab:ablation_teacher}
\begin{tabular}{@{}lccc@{}}
\toprule
\textcolor{tabSub}{\footnotesize Domain (avg@8, \%)} & \textcolor{tabSub}{\footnotesize Qwen3.8-Flash-Next traces} & \textcolor{tabSub}{\footnotesize Self traces} & \textcolor{tabSub}{\footnotesize $\Delta$} \\
\midrule
\domdot{domImgr}\,Image STEM & 60.6 & 63.4 & \dchip{p}{2.8} \\
\domdot{domImgp}\,Image perception & 77.2 & 77.7 & \dchip{p}{0.5} \\
\domdot{domVid}\,Video & 59.2 & 59.1 & \dchip{n}{0.1} \\
\bottomrule
\end{tabular}
\end{table}
\fi%
\paragraph{Choice of the trace generator.}
For each domain, we consider two generators: Qwen3.5-9B itself, which gives self traces, and the larger and stronger Qwen3.8-Flash-Next~\cite{qwen2026qwen38flashnext}.
A stronger generator can write better traces, but work on OPD shows that a stronger teacher is not always a better one, because the gain depends on the compatibility between the student and the teacher~\cite{acikgoz2026synergy,li2026rethinkingopd,ma2026mopd}.
Self traces, in contrast, stay close to the response distribution of Qwen3.5-9B, so SFT on them changes the reasoning style of the model less before the later stages.
We therefore train one SFT model of each domain on each trace source and compare the two (\ifdefined\ppmode Table~\ref{tab:ablation_teacher}\else\suppsec{A.4}\fi).
On both image domains, self traces give the better model, with a clear margin on image STEM, whereas on video, self traces give a slightly worse model than the Qwen3.8-Flash-Next traces.
We thus use self traces for image STEM and image perception and Qwen3.8-Flash-Next traces for video.
The final SFT set, \oursft, has 1.35M samples: the 994K image self traces (368K for image STEM and 626K for image perception) and one Qwen3.8-Flash-Next trace for each of the 357K video questions.

\paragraph{RL prompts.}
The RL sets keep only prompts with a rule-verifiable answer: 64,307 for image STEM, 67,915 for image perception and 60,805 for video.
The union of the three sets (193,027 prompts) forms \ourrl, the prompt set of MOPD and MixRL, where each prompt keeps its domain label.

\paragraph{Prompt template.}
All stages use the native Qwen3.5 chat template with thinking enabled and no system prompt.
The user turn holds the media and the question, followed by one format line for the answer type, \eg ``Put your final answer within \texttt{\textbackslash boxed\{\}}.'' for math.
We also remove the answer-format instructions of the source questions, so each prompt has exactly one format instruction.

\paragraph{Deduplication and decontamination.}
In the SFT question pool, we remove samples with near-identical media and the same normalized question across all domains.
We then decontaminate all SFT and RL sets against the evaluation benchmarks of Section~\ref{sec:setup}.
We remove a sample if it contains at least half of the word 13-grams of a benchmark question, if one of its images is within a perceptual-hash (pHash) distance of 4 of a benchmark image, or if its video shares a source video identifier or a frame hash with a benchmark video.
Finally, the SFT and RL sets are disjoint on the prompt and media hash.

\subsection{Reward}\label{sec:recipe_reward}

All RL stages, namely the domain teachers, the MixRL baseline, and MixRL on top of MOPD, share the same GSPO settings, and they score each response with one rule-based verifiable reward,
\begin{equation}
  R = 0.9\,R_\text{acc} + 0.1\,R_\text{fmt},
  \label{eq:reward}
\end{equation}
where $R_\text{acc}\in[0,1]$ is the accuracy term and $R_\text{fmt}$ is the format term.
The format term checks the thinking block first and the box second:
\begin{equation}
  R_\text{fmt} =
  \begin{cases}
    1   & \text{block valid, exactly one box,}\\
    0.5 & \text{block valid, not exactly one box,}\\
    0   & \text{block not valid.}
  \end{cases}
  \label{eq:fmt}
\end{equation}
Here a valid thinking block means that the response closes the block exactly once, with non-empty reasoning before \texttt{</think>} and a non-empty answer after it, and the chat template opens the block in the prompt, so the model only has to close it.
A box counts only in the answer part after \texttt{</think>}.
The accuracy term compares the boxed answer with the gold answer by the rule of its answer type, and it is zero unless $R_\text{fmt}=1$.%
\ifdefined\ppmode\space Appendix~B gives the matcher of each answer type.\fi

\subsection{Domain teachers}\label{sec:recipe_teachers}

\paragraph{GSPO teachers.}
We train the teacher $\pi_d$ of domain $d$ on its RL prompts with GSPO~\cite{zheng2025gspo}, directly from Qwen3.5-9B.
Other recipes use SFT as a warm-up for RL~\cite{guo2025deepseekr1,zhang2025openmmreasoner,lin2026mmvistareason}, so we also train a second set of teachers with the same settings from an SFT warm-up, which fine-tunes Qwen3.5-9B on the SFT set of the domain (Section~\ref{sec:recipe_data}).
For a prompt $x$, GSPO samples a group of $G$ responses $\{y_i\}_{i=1}^{G}$ from the rollout policy $\pi_{\theta_\text{old}}$ and maximizes a clipped objective on whole sequences:
\begin{equation}
  \mathcal{J}(\theta) = \mathbb{E}_{x}\Big[\frac{1}{G}\sum_{i=1}^{G}\min\big(s_i\hat{A}_i,\ \bar{s}_i\hat{A}_i\big)\Big],
  \label{eq:gspo}
\end{equation}
where $\mathbb{E}_{x}$ averages over the RL prompts $x$ (media and question) of the domain, $\pi_\theta$ is the policy under training with weights $\theta$, and $\pi_{\theta_\text{old}}$ is the rollout policy, whose fixed weights $\theta_\text{old}$ sampled the group.
The ratio $s_i=\big(\pi_\theta(y_i|x)/\pi_{\theta_\text{old}}(y_i|x)\big)^{1/|y_i|}$ measures, per token, how far the policy has moved on response $y_i$ of $|y_i|$ tokens, and $\bar{s}_i$ clips it to $[1-\epsilon_\text{l},1+\epsilon_\text{h}]$, so that the minimum removes any gain from a ratio outside this range.
The advantage $\hat{A}_i=(R_i-\mu)/\sigma$ is the reward $R_i$ of $y_i$ (Eq.~\ref{eq:reward}), normalized by the mean $\mu$ and the standard deviation $\sigma$ of the rewards in the group, and all tokens of a response therefore share one clipped ratio and one advantage.%
\ifdefined\ppmode\space We drop a prompt from the update when all of its responses are correct or all are wrong, because such a group gives no advantage signal.\fi

\subsection{Multi-Teacher On-Policy Distillation}\label{sec:recipe_mopd}

MOPD trains one student $\pi_\theta$ to match the teacher of each domain on its own samples.
Let $\mathcal{D}$ be the union of the RL prompt sets.
For each prompt $x\in\mathcal{D}$, the student samples one response $y\sim\pi_\theta(\cdot|x)$, and the frozen teacher $\pi_{r(x)}$ of the domain label $r(x)$ of the prompt scores the full sequence and returns, at each response position $t$, the log-probabilities of its top-$k$ token set $\mathcal{K}_t$.
The token loss at position $t$ is:
\begin{equation}
  \ell_t(\theta) = \sum_{v\in\mathcal{K}_t} \Big[\, p_\theta(v)\log\frac{p_\theta(v)}{q(v)} - p_\theta(v) + q(v) \Big],
  \label{eq:mopd_token}
\end{equation}
where $v$ runs over the tokens of $\mathcal{K}_t$, and $p_\theta(v)=\pi_\theta(v|x,y_{<t})$ and $q(v)=\pi_{r(x)}(v|x,y_{<t})$ are the probabilities that the student and the routed teacher give to $v$, given the prompt $x$ and the response tokens $y_{<t}$ before position $t$.
The first term is the reverse KL divergence restricted to $\mathcal{K}_t$, and the tail term $-p_\theta(v)+q(v)$ makes each summand non-negative and zero only at $p_\theta(v)=q(v)$, because neither distribution sums to one on $\mathcal{K}_t$.
\ifdefined\ppmode The batch loss is a token mean,
\begin{equation}
  \mathcal{L}(\theta) = \frac{\sum_{(x,y)\in\mathcal{B}} m(y) \sum_{t} \min\big(\ell_t(\theta),\, c\big)}{\sum_{(x,y)\in\mathcal{B}} m(y)\,|y|},
  \label{eq:supp_batch}
\end{equation}
where $\mathcal{B}$ is the batch of prompts $x$ with their student responses $y$, $t$ runs over the positions of $y$, $|y|$ is the token count of $y$, $c=\svClamp$ is the per-token clamp, which gives zero gradient above it, and the mask $m(y)$ is 0 if $y$ reaches the length cap of \svMopdResp{} tokens without an end token and 1 otherwise.
Without the mask, the teacher signal on a cut-off text would push the student toward text that no finished answer contains.
We differentiate the loss directly through $p_\theta$, with no task reward. Algorithm~\ref{alg:supp_mopd} gives one training step, and Appendix~E gives the routing, the asynchrony and the configuration.%
\begingroup\expandafter\let\expandafter\algorithm\csname algorithm*\endcsname\expandafter\let\expandafter\endalgorithm\csname endalgorithm*\endcsname\begin{algorithm}[!t]
\caption{One MOPD training step.}
\label{alg:supp_mopd}
\small
\begin{algorithmic}[1]
\Require student $\pi_\theta$, frozen teachers $\{\pi_d\}_{d\in\{\text{stem},\text{perc},\text{vid}\}}$, prompt union $\mathcal{D}$ with domain labels $r(x)$
\State Draw a batch $\mathcal{B}$ of 1008 prompts from $\mathcal{D}$
\ForAll{$x \in \mathcal{B}$} \Comment{student rollout, overlaps the update of the previous step}
\State Sample $y \sim \pi_{\theta'}(\cdot\mid x)$ at temperature 1, at most 16384 tokens \Comment{$\theta'$: weights at most 2 versions old}
\State Once $y$ ends, send $(x, y)$ to a replica of teacher $\pi_{r(x)}$
\State Teacher prefill: for each position $t$, the top-32 set $\mathcal{K}_t$ and $\log q(v)$, $v\in\mathcal{K}_t$, floored at $-10$
\State $m(y) \gets 0$ if $y$ hit the length cap without an end token, else $1$
\EndFor
\ForAll{$(x,y)\in\mathcal{B}$, positions $t$} \Comment{trainer}
\State $p_\theta(v) \gets \pi_\theta(v\mid x, y_{<t})$ on $\mathcal{K}_t$ (chunked over positions), $\log p_\theta(v)$ floored at $-10$ with the gradient of the unfloored value
\State $\ell_t \gets \sum_{v\in\mathcal{K}_t} \big[p_\theta(v)\log\frac{p_\theta(v)}{q(v)} - p_\theta(v) + q(v)\big]$
\State $\ell_t \gets \min(\ell_t, 10)$
\EndFor
\State $\mathcal{L} \gets \sum m(y)\sum_t \ell_t \,/\, \sum m(y)\,|y|$ \Comment{token mean over unmasked responses}
\State One AdamW update of $\theta$ on $\nabla_\theta\mathcal{L}$; sync the new weights to the rollout engines
\end{algorithmic}
\end{algorithm}
\endgroup
\else The training loss is the mean of $\min(\ell_t, c)$, with a constant clamp $c$, over the response tokens of a batch (\suppsec{E}), and we differentiate it directly through $p_\theta$, with no task reward.\fi{}
We initialize the student from Qwen3.5-9B, the model that all teachers come from, and we run MOPD once for each set of three teachers.

\subsection{MixRL on top of MOPD}\label{sec:recipe_mixrl}

In the common recipe, the RL prompts serve a single stage, MixRL from the base or SFT model, which has to merge the capabilities of all domains into one policy and to strengthen its reasoning at the same time.
These two roles compete: MixRL can harm individual capabilities~\cite{lin2026mmvistareason,xiong2026scoped}, and the MixSFT model has mainly learned a reasoning style, so it must learn all domains from a sparse reward at once while it also sharpens its reasoning.
In our recipe, MOPD assumes the first role: distillation merges the capabilities of all three teachers into one policy and already matches or outperforms the MixRL baseline (Section~\ref{sec:main_results}).
The MOPD student is therefore a stronger starting point for RL than the model from which MixRL usually starts.
At the same time, MOPD trains the student to match its teachers, so the teachers bound what the distillation loss can teach, whereas a reward-driven stage does not depend on any teacher and can move the student past this bound.
We therefore use the prompts once more, with MixRL on top of the MOPD student, to examine whether RL from this starting point further improves the model on the same data.
We initialize MixRL from the final MOPD checkpoint and train it on the prompt union \ourrl with the GSPO settings and the reward of the teachers, as for the MixRL baseline, which starts from MixSFT instead.
With the teachers trained directly from Qwen3.5-9B, the result is our main model, \oursmodel.

\section{Experiments}\label{sec:exp}

\begin{table*}[t]
\centering\footnotesize
\setlength{\tabcolsep}{4.6pt}
\renewcommand{\arraystretch}{1.05}
\caption{Results stage by stage (domain means and the mean of all 21 benchmarks, avg@8 accuracy, \%). The check marks show the training stages of each row: SFT, GSPO on the prompts of one domain (Domain RL), MOPD, and MixRL. $^\dagger$Three models, one per domain, each evaluated on the benchmarks of its own domain. \textbf{Bold} marks the best in each column, without the $^\dagger$ rows.}
\label{tab:stages}
\begin{tabular}{@{}l@{\hspace{14pt}}cccc@{\hspace{18pt}}ccc@{\hspace{18pt}}c@{}}
\toprule
 & \multicolumn{4}{c@{\hspace{18pt}}}{\textbf{Stages}} & \multicolumn{4}{c}{\textbf{Domain mean and overall mean, avg@8 (\%)}} \\
\cmidrule(r{18pt}){2-5}\cmidrule{6-9}
\textbf{Model} & SFT & Domain RL & MOPD & MixRL & \domtag{domImgr}{Image STEM} & \domtag{domImgp}{Image perception} & \domtag{domVid}{Video} & \textbf{Overall} \\
\midrule
Qwen3.5-9B &  &  &  &  & 70.0 & 78.9 & 53.4 & 69.4 \\
\midrule
\tgroup{9}{Mixed training on all domains} \\
MixRL &  &  &  & \ding{51} & 74.9 & 80.8 & 63.4 & 74.4 \\
MixSFT & \ding{51} &  &  &  & 63.4 & 77.7 & 58.8 & 67.7 \\
+ MixRL (MixRL baseline) & \ding{51} &  &  & \ding{51} & 73.8 & 80.4 & 62.1 & 73.5 \\
\midrule
\tgroup{9}{Domain teachers with the SFT warm-up} \\
SFT warm-ups$^\dagger$ & \ding{51} &  &  &  & 63.4 & 77.7 & 59.2 & 67.8 \\
+ GSPO (domain teachers)$^\dagger$ & \ding{51} & \ding{51} &  &  & 74.9 & 80.2 & 62.9 & 74.1 \\
+ MOPD & \ding{51} & \ding{51} & \ding{51} &  & 74.8 & 80.6 & 62.8 & 74.2 \\
+ MixRL & \ding{51} & \ding{51} & \ding{51} & \ding{51} & 75.3 & 81.2 & 63.7 & 74.8 \\
\midrule
\tgroup{9}{Domain teachers without the SFT warm-up} \\
GSPO (domain teachers)$^\dagger$ &  & \ding{51} &  &  & 76.7 & 81.2 & 63.4 & 75.2 \\
+ MOPD &  & \ding{51} & \ding{51} &  & 76.6 & 81.4 & 63.2 & 75.2 \\
\noalign{{\color{tabOurs}\hrule height \dimexpr\arraystretch\ht\strutbox+\arraystretch\dp\strutbox\relax}\vskip-\dimexpr\arraystretch\ht\strutbox+\arraystretch\dp\strutbox\relax}
\textbf{+ MixRL (\oursmodel)} &  & \ding{51} & \ding{51} & \ding{51} & \textbf{77.1} & \textbf{81.9} & \textbf{64.1} & \textbf{75.9} \\
\bottomrule
\end{tabular}
\end{table*}
\begin{table*}[t]
\centering
\footnotesize
\setlength{\tabcolsep}{0.9pt}
\renewcommand{\arraystretch}{0.86}
\caption{Results on all 21 benchmarks (avg@8 accuracy, \%). Each domain block starts with the domain mean, and ``Overall'' averages all 21 benchmarks. $\Delta$ is the change from Qwen3.5-9B. The mixed-training columns show MixRL directly from Qwen3.5-9B, MixSFT, and MixRL from MixSFT (the MixRL baseline). For the teachers with and without the SFT warm-up, we show the domain teachers, their MOPD student, and MixRL on top of that student. $^\star$ marks \oursmodel. In each row, \textbf{bold} marks the best among all columns except the teachers. The domain teachers are three models, one per domain (colored bar), each evaluated only on its own domain, so their ``Overall'' combines the three models.}
\label{tab:main_results}
\resizebox{\textwidth}{!}{%
\begin{tabular}{l>{\hspace{5pt}}c>{\hspace{3.4pt}}r@{\hspace{3pt}}l>{\hspace{3.4pt}}r@{\hspace{3pt}}l>{\hspace{3.4pt}}r@{\hspace{3pt}}l@{\hspace{2pt}\vrule width 0.4pt\hspace{3pt}}c@{\hspace{1pt}}>{\hspace{3.4pt}}r@{\hspace{3pt}}l>{\hspace{3.4pt}}r@{\hspace{3pt}}l>{\hspace{3.4pt}}r@{\hspace{3pt}}l@{\hspace{2pt}\vrule width 0.4pt\hspace{3pt}}c@{\hspace{1pt}}>{\hspace{3.4pt}}r@{\hspace{3pt}}l>{\hspace{3.4pt}}r@{\hspace{3pt}}l>{\hspace{3.4pt}}r@{\hspace{3pt}}l}
\toprule
 & \textbf{Base} & \multicolumn{6}{c@{\hspace{2pt}\vrule width 0.4pt\hspace{3pt}}}{\textbf{Mixed training}} & \multicolumn{7}{c@{\hspace{2pt}\vrule width 0.4pt\hspace{3pt}}}{\textbf{With the SFT warm-up}} & \multicolumn{7}{c}{\textbf{Without the SFT warm-up}} \\
\cmidrule(l{5pt}r){3-8}\cmidrule(l{5pt}r){9-15}\cmidrule(l{5pt}){16-22}
\textbf{Benchmark} & \textbf{Qwen3.5-9B} & \multicolumn{2}{c}{\hspace{3.4pt}\textbf{MixRL}} & \multicolumn{2}{c}{\hspace{3.4pt}\textbf{MixSFT}} & \multicolumn{2}{c@{\hspace{2pt}\vrule width 0.4pt\hspace{3pt}}}{\hspace{3.4pt}\textbf{+ MixRL}} &  & \multicolumn{2}{c}{\hspace{3.4pt}\textbf{Teachers}} & \multicolumn{2}{c}{\hspace{3.4pt}\textbf{MOPD}} & \multicolumn{2}{c@{\hspace{2pt}\vrule width 0.4pt\hspace{3pt}}}{\hspace{3.4pt}\textbf{+ MixRL}} &  & \multicolumn{2}{c}{\hspace{3.4pt}\textbf{Teachers}} & \multicolumn{2}{c}{\hspace{3.4pt}\textbf{MOPD}} & \multicolumn{2}{c}{\hspace{3.4pt}\textbf{+ MixRL$^\star$}} \\
 & avg@8 & \multicolumn{1}{c}{\hspace{3.4pt}avg@8} & \makebox[2.45em][r]{$\Delta$} & \multicolumn{1}{c}{\hspace{3.4pt}avg@8} & \makebox[2.45em][r]{$\Delta$} & \multicolumn{1}{c}{\hspace{3.4pt}avg@8} & \makebox[2.45em][r]{$\Delta$} &  & \multicolumn{1}{c}{\hspace{3.4pt}avg@8} & \makebox[2.45em][r]{$\Delta$} & \multicolumn{1}{c}{\hspace{3.4pt}avg@8} & \makebox[2.45em][r]{$\Delta$} & \multicolumn{1}{c}{\hspace{3.4pt}avg@8} & \makebox[2.45em][r]{$\Delta$} &  & \multicolumn{1}{c}{\hspace{3.4pt}avg@8} & \makebox[2.45em][r]{$\Delta$} & \multicolumn{1}{c}{\hspace{3.4pt}avg@8} & \makebox[2.45em][r]{$\Delta$} & \multicolumn{1}{c}{\hspace{3.4pt}avg@8} & \makebox[2.45em][r]{$\Delta$} \\
\midrule
\domtag{domImgr}{Image STEM} \textbf{mean} & 70.0 & 74.9 & \dchip{p}{4.9} & 63.4 & \dchip{n}{6.6} & 73.8 & \dchip{p}{3.8} & \cellcolor{domImgr} & 74.9 & \dchip{p}{4.9} & 74.8 & \dchip{p}{4.8} & 75.3 & \dchip{p}{5.3} & \cellcolor{domImgr} & 76.7 & \dchip{p}{6.7} & 76.6 & \dchip{p}{6.6} & \textbf{77.1} & \dchip{p}{7.1} \\
DynaMath & 74.4 & 76.1 & \dchip{p}{1.7} & 72.1 & \dchip{n}{2.3} & 77.0 & \dchip{p}{2.6} & \cellcolor{domImgr} & 78.2 & \dchip{p}{3.8} & 78.1 & \dchip{p}{3.7} & 78.7 & \dchip{p}{4.3} & \cellcolor{domImgr} & 79.2 & \dchip{p}{4.8} & 79.5 & \dchip{p}{5.1} & \textbf{80.0} & \dchip{p}{5.6} \\
EMMA & 43.0 & 54.0 & \dchip{p}{11.0} & 43.0 & \dchip{z}{0.0} & 50.4 & \dchip{p}{7.4} & \cellcolor{domImgr} & 54.0 & \dchip{p}{11.0} & 53.5 & \dchip{p}{10.5} & 53.5 & \dchip{p}{10.5} & \cellcolor{domImgr} & 57.0 & \dchip{p}{14.0} & 57.0 & \dchip{p}{14.0} & \textbf{57.5} & \dchip{p}{14.5} \\
LogicVista & 67.6 & 75.1 & \dchip{p}{7.5} & 66.4 & \dchip{n}{1.2} & 72.4 & \dchip{p}{4.8} & \cellcolor{domImgr} & 72.6 & \dchip{p}{5.0} & 71.7 & \dchip{p}{4.1} & 73.8 & \dchip{p}{6.2} & \cellcolor{domImgr} & 78.2 & \dchip{p}{10.6} & 78.5 & \dchip{p}{10.9} & \textbf{79.6} & \dchip{p}{12.0} \\
MathVerse & 80.4 & 83.6 & \dchip{p}{3.2} & 76.1 & \dchip{n}{4.3} & 82.8 & \dchip{p}{2.4} & \cellcolor{domImgr} & 83.3 & \dchip{p}{2.9} & 83.1 & \dchip{p}{2.7} & 84.1 & \dchip{p}{3.7} & \cellcolor{domImgr} & 84.4 & \dchip{p}{4.0} & 84.2 & \dchip{p}{3.8} & \textbf{84.6} & \dchip{p}{4.2} \\
MathVista & 83.0 & 85.9 & \dchip{p}{2.9} & 81.5 & \dchip{n}{1.5} & 85.6 & \dchip{p}{2.6} & \cellcolor{domImgr} & 85.5 & \dchip{p}{2.5} & 86.2 & \dchip{p}{3.2} & 85.5 & \dchip{p}{2.5} & \cellcolor{domImgr} & 86.9 & \dchip{p}{3.9} & 86.7 & \dchip{p}{3.7} & \textbf{87.1} & \dchip{p}{4.1} \\
MMMU & 71.7 & 75.0 & \dchip{p}{3.3} & 71.2 & \dchip{n}{0.5} & 75.3 & \dchip{p}{3.6} & \cellcolor{domImgr} & 75.5 & \dchip{p}{3.8} & 75.9 & \dchip{p}{4.2} & \textbf{76.4} & \dchip{p}{4.7} & \cellcolor{domImgr} & 76.1 & \dchip{p}{4.4} & 75.9 & \dchip{p}{4.2} & 76.0 & \dchip{p}{4.3} \\
MMMU-Pro & 64.8 & 68.0 & \dchip{p}{3.2} & 62.4 & \dchip{n}{2.4} & 67.4 & \dchip{p}{2.6} & \cellcolor{domImgr} & 68.5 & \dchip{p}{3.7} & 68.7 & \dchip{p}{3.9} & 68.8 & \dchip{p}{4.0} & \cellcolor{domImgr} & 69.9 & \dchip{p}{5.1} & 69.2 & \dchip{p}{4.4} & \textbf{69.7} & \dchip{p}{4.9} \\
We-Math & 74.8 & 81.0 & \dchip{p}{6.2} & 34.3 & \dchip{n}{40.5} & 79.2 & \dchip{p}{4.4} & \cellcolor{domImgr}\multirow{-9}{*}{\rotatebox{90}{\fontsize{5.6}{6}\selectfont\bfseries\textcolor{white}{Image STEM models}}} & 81.8 & \dchip{p}{7.0} & 81.2 & \dchip{p}{6.4} & 81.9 & \dchip{p}{7.1} & \cellcolor{domImgr}\multirow{-9}{*}{\rotatebox{90}{\fontsize{5.6}{6}\selectfont\bfseries\textcolor{white}{Image STEM models}}} & 81.7 & \dchip{p}{6.9} & 81.4 & \dchip{p}{6.6} & \textbf{82.5} & \dchip{p}{7.7} \\
\midrule
\domtag{domImgp}{Image perception} \textbf{mean} & 78.9 & 80.8 & \dchip{p}{1.9} & 77.7 & \dchip{n}{1.2} & 80.4 & \dchip{p}{1.5} & \cellcolor{domImgp} & 80.2 & \dchip{p}{1.3} & 80.6 & \dchip{p}{1.7} & 81.2 & \dchip{p}{2.3} & \cellcolor{domImgp} & 81.2 & \dchip{p}{2.3} & 81.4 & \dchip{p}{2.5} & \textbf{81.9} & \dchip{p}{3.0} \\
AI2D & 86.8 & 89.7 & \dchip{p}{2.9} & 86.2 & \dchip{n}{0.6} & 89.2 & \dchip{p}{2.4} & \cellcolor{domImgp} & 89.2 & \dchip{p}{2.4} & 89.5 & \dchip{p}{2.7} & 90.3 & \dchip{p}{3.5} & \cellcolor{domImgp} & 89.4 & \dchip{p}{2.6} & 89.9 & \dchip{p}{3.1} & \textbf{92.1} & \dchip{p}{5.3} \\
CountBenchQA & 97.1 & 97.3 & \dchip{p}{0.2} & 95.0 & \dchip{n}{2.1} & 96.2 & \dchip{n}{0.9} & \cellcolor{domImgp} & 96.3 & \dchip{n}{0.8} & 97.2 & \dchip{p}{0.1} & 97.5 & \dchip{p}{0.4} & \cellcolor{domImgp} & 97.6 & \dchip{p}{0.5} & 97.7 & \dchip{p}{0.6} & \textbf{98.4} & \dchip{p}{1.3} \\
CV-Bench & 86.4 & \textbf{89.4} & \dchip{p}{3.0} & 86.8 & \dchip{p}{0.4} & 89.2 & \dchip{p}{2.8} & \cellcolor{domImgp} & 88.6 & \dchip{p}{2.2} & 88.7 & \dchip{p}{2.3} & 88.6 & \dchip{p}{2.2} & \cellcolor{domImgp} & 89.6 & \dchip{p}{3.2} & \textbf{89.4} & \dchip{p}{3.0} & 89.1 & \dchip{p}{2.7} \\
HallusionBench & 78.4 & 78.4 & \dchip{z}{0.0} & 76.2 & \dchip{n}{2.2} & 78.5 & \dchip{p}{0.1} & \cellcolor{domImgp} & 77.4 & \dchip{n}{1.0} & 78.2 & \dchip{n}{0.2} & 77.2 & \dchip{n}{1.2} & \cellcolor{domImgp} & 78.3 & \dchip{n}{0.1} & \textbf{79.1} & \dchip{p}{0.7} & 78.9 & \dchip{p}{0.5} \\
MMStar & 75.3 & 77.9 & \dchip{p}{2.6} & 74.1 & \dchip{n}{1.2} & 77.2 & \dchip{p}{1.9} & \cellcolor{domImgp} & 77.0 & \dchip{p}{1.7} & 78.2 & \dchip{p}{2.9} & 78.4 & \dchip{p}{3.1} & \cellcolor{domImgp} & 78.6 & \dchip{p}{3.3} & 79.0 & \dchip{p}{3.7} & \textbf{79.3} & \dchip{p}{4.0} \\
OCRBench & 87.1 & 89.6 & \dchip{p}{2.5} & 85.6 & \dchip{n}{1.5} & 89.5 & \dchip{p}{2.4} & \cellcolor{domImgp} & 89.8 & \dchip{p}{2.7} & 89.6 & \dchip{p}{2.5} & 89.7 & \dchip{p}{2.6} & \cellcolor{domImgp} & 89.5 & \dchip{p}{2.4} & 89.3 & \dchip{p}{2.2} & \textbf{90.0} & \dchip{p}{2.9} \\
RealWorldQA & 75.4 & 76.1 & \dchip{p}{0.7} & 74.1 & \dchip{n}{1.3} & 76.7 & \dchip{p}{1.3} & \cellcolor{domImgp} & 77.0 & \dchip{p}{1.6} & 75.9 & \dchip{p}{0.5} & \textbf{78.9} & \dchip{p}{3.5} & \cellcolor{domImgp} & 78.5 & \dchip{p}{3.1} & 78.6 & \dchip{p}{3.2} & 78.6 & \dchip{p}{3.2} \\
SimpleVQA & 44.7 & 48.2 & \dchip{p}{3.5} & 43.5 & \dchip{n}{1.2} & 46.2 & \dchip{p}{1.5} & \cellcolor{domImgp}\multirow{-9}{*}{\rotatebox{90}{\fontsize{5.6}{6}\selectfont\bfseries\textcolor{white}{Image perception models}}} & 46.1 & \dchip{p}{1.4} & 47.6 & \dchip{p}{2.9} & 48.8 & \dchip{p}{4.1} & \cellcolor{domImgp}\multirow{-9}{*}{\rotatebox{90}{\fontsize{5.6}{6}\selectfont\bfseries\textcolor{white}{Image perception models}}} & 48.1 & \dchip{p}{3.4} & 48.4 & \dchip{p}{3.7} & \textbf{49.1} & \dchip{p}{4.4} \\
\midrule
\domtag{domVid}{Video} \textbf{mean} & 53.4 & 63.4 & \dchip{p}{10.0} & 58.8 & \dchip{p}{5.4} & 62.1 & \dchip{p}{8.7} & \cellcolor{domVid} & 62.9 & \dchip{p}{9.5} & 62.8 & \dchip{p}{9.4} & 63.7 & \dchip{p}{10.3} & \cellcolor{domVid} & 63.4 & \dchip{p}{10.0} & 63.2 & \dchip{p}{9.8} & \textbf{64.1} & \dchip{p}{10.7} \\
LongVideoBench & 51.0 & \textbf{63.4} & \dchip{p}{12.4} & 58.0 & \dchip{p}{7.0} & 61.0 & \dchip{p}{10.0} & \cellcolor{domVid} & 63.6 & \dchip{p}{12.6} & 63.1 & \dchip{p}{12.1} & 62.5 & \dchip{p}{11.5} & \cellcolor{domVid} & 63.0 & \dchip{p}{12.0} & 62.4 & \dchip{p}{11.4} & 63.3 & \dchip{p}{12.3} \\
TempCompass & 74.9 & \textbf{76.8} & \dchip{p}{1.9} & 74.0 & \dchip{n}{0.9} & 75.2 & \dchip{p}{0.3} & \cellcolor{domVid} & 74.9 & \dchip{z}{0.0} & 75.4 & \dchip{p}{0.5} & 75.9 & \dchip{p}{1.0} & \cellcolor{domVid} & 75.8 & \dchip{p}{0.9} & \textbf{76.8} & \dchip{p}{1.9} & 76.2 & \dchip{p}{1.3} \\
Video-Holmes & 42.8 & 51.7 & \dchip{p}{8.9} & 46.9 & \dchip{p}{4.1} & 51.0 & \dchip{p}{8.2} & \cellcolor{domVid} & 52.4 & \dchip{p}{9.6} & 51.7 & \dchip{p}{8.9} & 52.9 & \dchip{p}{10.1} & \cellcolor{domVid} & 52.0 & \dchip{p}{9.2} & 51.3 & \dchip{p}{8.5} & \textbf{53.1} & \dchip{p}{10.3} \\
Video-MME & 58.5 & \textbf{70.7} & \dchip{p}{12.2} & 66.4 & \dchip{p}{7.9} & 69.2 & \dchip{p}{10.7} & \cellcolor{domVid} & 69.5 & \dchip{p}{11.0} & 70.0 & \dchip{p}{11.5} & 69.9 & \dchip{p}{11.4} & \cellcolor{domVid} & 70.0 & \dchip{p}{11.5} & 70.2 & \dchip{p}{11.7} & 70.0 & \dchip{p}{11.5} \\
VSI-Bench & 39.8 & 54.5 & \dchip{p}{14.7} & 48.4 & \dchip{p}{8.6} & 54.0 & \dchip{p}{14.2} & \cellcolor{domVid}\multirow{-6}{*}{\rotatebox{90}{\fontsize{5.6}{6}\selectfont\bfseries\textcolor{white}{Video models}}} & 54.2 & \dchip{p}{14.4} & 54.1 & \dchip{p}{14.3} & 57.1 & \dchip{p}{17.3} & \cellcolor{domVid}\multirow{-6}{*}{\rotatebox{90}{\fontsize{5.6}{6}\selectfont\bfseries\textcolor{white}{Video models}}} & 56.2 & \dchip{p}{16.4} & 55.2 & \dchip{p}{15.4} & \textbf{58.0} & \dchip{p}{18.2} \\
\midrule
\domtag{figInk}{Overall} \textbf{mean} & 69.4 & 74.4 & \dchip{p}{5.0} & 67.7 & \dchip{n}{1.7} & 73.5 & \dchip{p}{4.1} & \cellcolor{figInk} & 74.1 & \dchip{p}{4.7} & 74.2 & \dchip{p}{4.8} & 74.8 & \dchip{p}{5.4} & \cellcolor{figInk} & 75.2 & \dchip{p}{5.8} & 75.2 & \dchip{p}{5.8} & \textbf{75.9} & \dchip{p}{6.5} \\
\bottomrule
\end{tabular}}
\end{table*}

\subsection{Setup}\label{sec:setup}

\paragraph{Models.}
We use Qwen3.5-9B~\cite{qwen2026qwen35} as the base model of every run, and we train SFT with lmms-engine~\cite{lmmsengine2025} and GSPO and MOPD with verl~\cite{sheng2025hybridflow}, with vLLM~\cite{kwon2023vllm} for the rollout.%
\ifdefined\ppmode\space Appendix~D gives all training settings.\fi

\paragraph{Benchmarks.}
We evaluate on 21 benchmarks in three domains, and for image STEM reasoning, we use DynaMath~\cite{zou2025dynamath}, EMMA~\cite{hao2025emma}, LogicVista~\cite{xiao2024logicvista}, MathVerse~\cite{zhang2024mathverse}, MathVista~\cite{lu2024mathvista}, MMMU~\cite{yue2024mmmu}, MMMU-Pro~\cite{yue2025mmmupro} and We-Math~\cite{qiao2024wemath}.
For image perception reasoning, we use AI2D~\cite{kembhavi2016ai2d}, CountBenchQA~\cite{paiss2023countbench,beyer2024paligemma}, CV-Bench~\cite{tong2024cambrian}, HallusionBench~\cite{guan2024hallusionbench}, MMStar~\cite{chen2024mmstar}, OCRBench~\cite{liu2024ocrbench}, RealWorldQA~\cite{xai2024realworldqa} and SimpleVQA~\cite{cheng2025simplevqa}.
For video reasoning, we use LongVideoBench~\cite{wu2024longvideobench}, TempCompass~\cite{liu2024tempcompass}, Video-Holmes~\cite{cheng2025videoholmes}, Video-MME~\cite{fu2025videomme} and VSI-Bench~\cite{yang2025thinkinginspace}.
In addition, all training data is decontaminated against these benchmarks (Section~\ref{sec:recipe_data}).

\paragraph{Evaluation.}
We evaluate all models with one protocol in lmms-eval~\cite{zhang2024lmmseval}, with thinking enabled and a cap of 32768 tokens per response, thinking included.
We sample eight responses per question (temperature 0.6, top-$p$ 0.95, top-$k$ 20) and report the mean accuracy over the eight samples (avg@8).
Images are resized to at most 1,605,632 pixels, and each video frame has at most 65,536 pixels.
Videos shorter than 150 seconds are sampled at 1 fps with at least 4 frames, and from longer videos we sample 150 evenly spaced frames.%
\ifdefined\ppmode\space Appendix~C gives the dataset, split and metric of each benchmark and the judge of SimpleVQA.\fi

\paragraph{Compared models.}
We compare two families of recipes on the same data.
The mixed-training recipes train one model on all domains with no teachers: MixSFT trains Qwen3.5-9B on the union of the three domain SFT sets (Section~\ref{sec:recipe_data}), the MixRL baseline runs GSPO from the MixSFT checkpoint on the same prompt union as MOPD, with the same GSPO settings as the domain teachers, and we also run the same MixRL directly from Qwen3.5-9B.
The multi-teacher recipes distill one set of domain teachers into one student with MOPD, where the teachers are trained with GSPO directly from Qwen3.5-9B or after an SFT warm-up.
For each set of teachers, MixRL on top of MOPD uses the same prompts and the same GSPO settings as the MixRL baseline (Section~\ref{sec:recipe_mixrl}).
The SFT warm-ups and the domain teachers are three models each, one per domain, and we evaluate each of them only on the benchmarks of its own domain, whereas we evaluate all other models on all 21 benchmarks.

\subsection{Main results}\label{sec:main_results}

\ifdefined\ppmode\else\begin{table*}[t]
\centering
\fontsize{6.0}{7.2}\selectfont
\setlength{\tabcolsep}{0.6pt}
\renewcommand{\arraystretch}{1.0}
\caption{Comparison with published results on 16 image benchmarks (left) and 5 video benchmarks (right), in accuracy (\%). The results of \oursmodel{} are avg@8. Best results are \textbf{bold} and the second-best are \underline{underlined}. OpenMMReasoner, InternVL3.5 and MMFineReason use the values of MMVistaReason~\cite{lin2026mmvistareason}, and Video-R1 on Video-Holmes that of the Video-Holmes paper~\cite{cheng2025videoholmes}. On EMMA and We-Math, OneThinker uses the values of~\cite{kulkarni2026videoflair,wu2026seeing}, OpenMMReasoner those of~\cite{wang2026ane} and~\cite{li2026verievol}, and InternVL3.5 on We-Math its own.}
\label{tab:compare}
{\renewcommand{\arraystretch}{0.92}%
\begin{tabular}[t]{@{}l>{\hspace{2pt}}cccccccc>{\hspace{2pt}}cccccccc@{}}
\toprule
 & \multicolumn{8}{c}{\domtag{domImgr}{Image STEM}} & \multicolumn{8}{c}{\domtag{domImgp}{Image perception}} \\
\cmidrule(l{2pt}r{1pt}){2-9}\cmidrule(l{2pt}r{1pt}){10-17}
\textbf{Model} & \shortstack{Dyna\\Math} & \shortstack{EMMA} & \shortstack{Logic\\Vista} & \shortstack{Math\\Verse} & \shortstack{Math\\Vista} & \shortstack{MMMU} & \shortstack{MMMU-\\Pro} & \shortstack{We-\\Math} & \shortstack{AI2D} & \shortstack{Count\\Bench\\QA} & \shortstack{CV-\\Bench} & \shortstack{Hallu-\\sion\\Bench} & \shortstack{MM\\Star} & \shortstack{OCR\\Bench} & \shortstack{Real\\World\\QA} & \shortstack{Simple\\VQA} \\
\midrule
OneThinker-8B~\cite{feng2025onethinker} & -- & 10.9 & -- & 64.3 & 77.6 & 70.6 & -- & 54.6 & 85.2 & -- & -- & -- & 70.6 & -- & -- & -- \\
Video-HopChain-8B~\cite{quang2026videohopchain} & -- & -- & -- & -- & -- & -- & -- & -- & 82.8 & -- & -- & -- & 67.5 & -- & 67.3 & -- \\
\midrule
OpenMMReasoner-7B~\cite{zhang2025openmmreasoner} & -- & \underline{24.5} & 51.9 & 63.7 & 79.3 & 57.6 & -- & 53.8 & -- & 81.9 & -- & -- & -- & -- & 68.8 & -- \\
InternVL3.5-30B-A3B~\cite{wang2025internvl35} & -- & -- & 53.9 & 61.8 & 75.6 & 69.4 & -- & 48.4 & -- & 84.7 & -- & -- & -- & -- & 72.3 & -- \\
MMFineReason-8B~\cite{lin2026mmfinereason} & -- & -- & 68.5 & 80.4 & 82.1 & 71.2 & -- & -- & -- & 88.1 & -- & -- & -- & -- & 75.2 & -- \\
MMVistaReason-9B~\cite{lin2026mmvistareason} & -- & -- & 68.5 & \underline{82.4} & 83.6 & 73.6 & -- & -- & -- & \underline{92.8} & -- & -- & -- & -- & 77.4 & -- \\
\midrule
HopChain-35B-A3B~\cite{wang2026hopchain} & -- & -- & \underline{75.6} & -- & \underline{85.0} & \textbf{78.3} & \textbf{70.6} & -- & \underline{91.3} & -- & -- & \underline{66.5} & \underline{78.6} & -- & \textbf{79.4} & -- \\
STEP3-VL-10B~\cite{stepfun2026step3vl} & -- & -- & 66.9 & -- & 84.0 & \underline{78.1} & 64.1 & 73.0 & 89.4 & 88.8 & \underline{83.5} & 64.9 & 77.5 & 86.8 & 74.4 & \underline{53.1} \\
MM-OPD-9B~\cite{tong2026mmopd} & \underline{77.7} & -- & 73.4 & -- & 84.1 & 75.8 & 64.7 & \underline{76.5} & -- & -- & -- & -- & 75.9 & \underline{87.0} & 77.5 & \textbf{53.2} \\
\midrule
\rowcolor{tabOurs}\textbf{\oursmodel} & \textbf{80.0} & \textbf{57.5} & \textbf{79.6} & \textbf{84.6} & \textbf{87.1} & 76.0 & \underline{69.7} & \textbf{82.5} & \textbf{92.1} & \textbf{98.4} & \textbf{89.1} & \textbf{78.9} & \textbf{79.3} & \textbf{90.0} & \underline{78.6} & 49.1 \\
\bottomrule
\end{tabular}}%
\hfill
{\renewcommand{\arraystretch}{0.92}%
\begin{tabular}[t]{@{}l>{\hspace{2pt}}ccccc@{}}
\toprule
 & \multicolumn{5}{c}{\domtag{domVid}{Video}} \\
\cmidrule(l{2pt}r{1pt}){2-6}
\textbf{Model} & \shortstack{Long\\Video\\Bench} & \shortstack{Temp\\Compass} & \shortstack{Video-\\Holmes} & \shortstack{Video-\\MME} & \shortstack{VSI-\\Bench} \\
\midrule
Video-R1-7B~\cite{feng2025videor1} & -- & 73.2 & 36.5 & 61.4 & 37.1 \\
Video-RTS-7B~\cite{wang2025videorts} & 56.6 & -- & 40.7 & 63.0 & -- \\
Video-KTR-7B~\cite{wang2026videoktr} & -- & 73.5 & 42.7 & 62.5 & -- \\
Video-o3-7B~\cite{zeng2026videoo3} & 60.5 & -- & 46.5 & 66.5 & -- \\
\midrule
OneThinker-8B~\cite{feng2025onethinker} & 61.7 & -- & 48.7 & 66.5 & -- \\
Video-HopChain-8B~\cite{quang2026videohopchain} & -- & -- & 48.6 & 69.2 & -- \\
\midrule
Video-MOPD-8B~\cite{qin2026videomopd} & -- & \textbf{81.9} & \textbf{54.7} & \textbf{72.7} & -- \\
Latent-OPD-9B~\cite{shen2026latentopd} & -- & 70.6 & -- & 66.5 & \underline{54.9} \\
RLER-7B~\cite{yang2026rler} & \underline{63.0} & \underline{76.2} & -- & 68.5 & 43.3 \\
\midrule
\rowcolor{tabOurs}\textbf{\oursmodel} & \textbf{63.3} & \underline{76.2} & \underline{53.1} & \underline{70.0} & \textbf{58.0} \\
\bottomrule
\end{tabular}}%
\end{table*}
\fi
\ifdefined\ppmode\begin{figure*}[t]
  \centering
  \includegraphics[width=\textwidth]{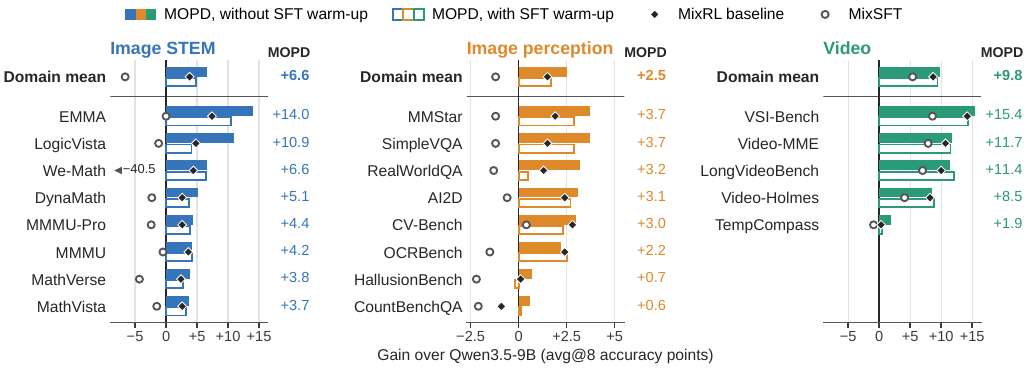}
  \caption{Gain over Qwen3.5-9B on each benchmark (avg@8 accuracy, points), with the domain mean first. Solid bars show MOPD of the teachers trained directly from Qwen3.5-9B (value on the right, which also sorts the benchmarks), and outlined bars show MOPD of the teachers with the SFT warm-up. An arrow with a value marks a gain outside the axis range. Both MOPD students improve on most benchmarks, and the MixRL baseline stays below them on most benchmarks.}
  \label{fig:gains}
\end{figure*}
\begin{figure*}[t]
  \centering
  \includegraphics[width=\textwidth]{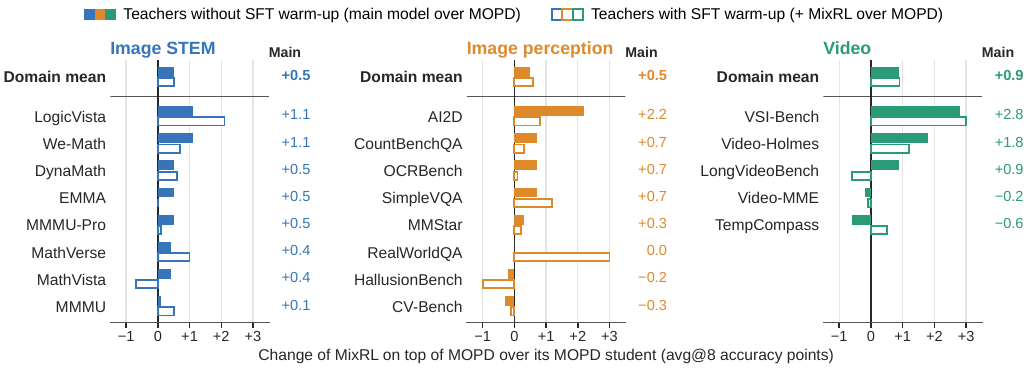}
  \caption{Change of MixRL on top of MOPD over its MOPD student on each benchmark (avg@8 accuracy, points), with the domain mean first. Solid bars show \oursmodel over the MOPD student of the teachers trained directly from Qwen3.5-9B (value on the right, which also sorts the benchmarks), and outlined bars show MixRL on top of the MOPD student of the teachers with the SFT warm-up. For both students, MixRL raises most benchmarks, and the few drops are small.}
  \label{fig:gains_mixrl}
\end{figure*}\fi%
\ifdefined\ppmode\fi%

Table~\ref{tab:stages} gives the domain means of every stage, and Table~\ref{tab:main_results} gives the results on all 21 benchmarks.%
\ifdefined\ppmode\space Figure~\ref{fig:gains} shows the gain of the two MOPD students, of MixSFT and of the MixRL baseline over Qwen3.5-9B on each benchmark.\fi

\paragraph{MOPD against the domain teachers.}
For both sets of teachers, the MOPD student stays close to each teacher on the domain of that teacher and matches the three teachers on average (Table~\ref{tab:stages}).
One student therefore keeps the skills of three domain teachers, and it does so for two different sets of teachers.

\paragraph{MOPD against mixed training.}
On the same prompts and with the same start model, MOPD is above mixed training: the MOPD student of the teachers trained directly from Qwen3.5-9B is above MixRL from Qwen3.5-9B on average and on most benchmarks (Table~\ref{tab:main_results}).
It is also above the MixRL baseline from MixSFT on every domain, which is consistent with earlier findings for MOPD on text~\cite{ma2026mopd} and on images~\cite{lin2026mmvistareason,xiong2026scoped}.
Likewise, the teachers with the SFT warm-up use the same SFT data as the MixRL baseline, so that only the training route differs, and their MOPD student is above the baseline on every domain and on most benchmarks.

\paragraph{Teachers with and without the SFT warm-up.}\label{sec:sft_warmup}
In our setting, the teachers trained directly from Qwen3.5-9B are slightly ahead of the teachers with the SFT warm-up on all three domains and on most benchmarks, and MOPD carries most of this difference to its student (Tables~\ref{tab:stages} and~\ref{tab:main_results}).
We therefore establish MOPD of the teachers trained directly from Qwen3.5-9B as our recipe.

\paragraph{MixRL on top of MOPD.}
Although MixRL trails MOPD on average as a separate route, the same prompts still carry the student further when MixRL starts from the MOPD student instead of from the base model.
On top of the MOPD student of our recipe, MixRL raises the overall mean and all three domain means, and this model, which has the highest overall mean of all models in Table~\ref{tab:stages}, is our main model, \oursmodel.
On top of the MOPD student of the teachers with the SFT warm-up, MixRL also raises these means, and this model is above all mixed-training models on every domain.
In both cases, the largest gain is on video, and MixRL on top of MOPD shows no collapse.%
\ifdefined\ppmode\space Per benchmark, MixRL on top of MOPD raises most of the 21 benchmarks for both students, and the few drops are small (Figure~\ref{fig:gains_mixrl}).\fi

\paragraph{Comparison with reasoning VLMs.}
Table~\ref{tab:compare} places our main model next to recent reasoning VLMs on our 21 benchmarks.
\oursmodel has the best or second-best result on almost all of these benchmarks, with best results in all three domains, which shows that the recipe gives one 9B model that is competitive across image STEM, image perception and video.
On average over the benchmarks that each row shares with us, it is above MMVistaReason and its image baselines, above OneThinker and Video-HopChain on both image and video, and above STEP3-VL-10B and MM-OPD on images.
On images, it is also above the larger HopChain-35B-A3B on average, whereas on video, it stays below Video-MOPD, which is built specifically for video understanding.

\section{Analysis}\label{sec:analysis_mixrl}

\begin{figure*}[t]
  \centering
  \includegraphics[width=0.9\textwidth]{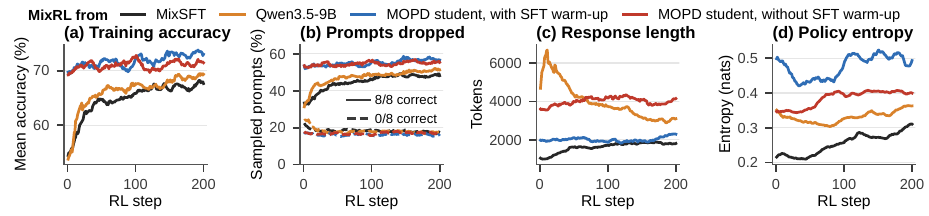}
  \caption{MixRL in the two mixed-training runs, from MixSFT (the MixRL baseline) and from Qwen3.5-9B, and from the two MOPD students, of the teachers with and without the SFT warm-up (the latter gives \oursmodel). (a)~Mean accuracy of the responses on the sampled training prompts, including the prompts that the group filter drops. (b)~Share of the sampled prompts that the group filter drops: prompts with all eight responses correct (8/8, solid) and with no correct response (0/8, dashed). (c)~Mean response length. (d)~Mean token entropy of the policy. Training metrics are 10-step moving averages.}
  \label{fig:mixrl_dynamics}
\end{figure*}

In this section, we analyze why MixRL on top of MOPD improves the student, first through the training dynamics of MixRL and then through the change in the reasoning of the model.

\paragraph{Training dynamics.}
In both mixed-training runs, the training accuracy rises during the run and the response length changes strongly, upward from MixSFT and downward from Qwen3.5-9B.
In contrast, both MOPD students start MixRL with a training accuracy that these runs only approach at the end of the shown steps, and change this length much less (Figure~\ref{fig:mixrl_dynamics}(a) and (c)).
Much of both mixed-training runs thus builds skills that the domain teachers have and that MOPD already transfers to the student.
The group filter drops a prompt when all of its responses are correct or all are wrong, because such a group gives no advantage signal.
From the first step, both students solve more than half of the sampled prompts in all eight responses and start with fewer prompts with no correct response than the mixed-training runs (Figure~\ref{fig:mixrl_dynamics}(b)), so the prompts that remain after the filter give their learning signal at a higher level of ability.
The entropy of the policy also differs (Figure~\ref{fig:mixrl_dynamics}(d)): MixRL from MixSFT starts with a low entropy, whereas the other three runs start with a higher entropy and keep it above this run, and no run shows a sign of entropy collapse.
Since RL mainly sharpens behavior that its start model already shows~\cite{yue2025rlbeyondbase,gandhi2025cognitive}, the start of MixRL decides how far the same prompts can carry the model.

\ifdefined\ppmode\begin{table*}[t]
\centering\footnotesize
\setlength{\tabcolsep}{3.1pt}
\renewcommand{\arraystretch}{1.05}
\caption{Reasoning behavior on the benchmarks (mean over the benchmarks of each domain). Len: thinking characters per response (K). H/E: thinking length on hard over easy questions. Chk: self-checks per 1K thinking characters. Loop, Hdg: responses with a loop or a hedge (\%). TS: cited timestamps per 1K thinking characters. $^\dagger$One model per domain.}
\label{tab:behavior}
\begin{tabular}{@{}l@{\hspace{8pt}}ccccc@{\hspace{10pt}}ccccc@{\hspace{10pt}}cccccc@{}}
\toprule
 & \multicolumn{5}{c@{\hspace{10pt}}}{\domtag{domImgr}{Image STEM}} & \multicolumn{5}{c@{\hspace{10pt}}}{\domtag{domImgp}{Image perception}} & \multicolumn{6}{c}{\domtag{domVid}{Video}} \\
\cmidrule(r{10pt}){2-6}\cmidrule(r{10pt}){7-11}\cmidrule{12-17}
\textbf{Model} & Len & H/E & Chk & Loop & Hdg & Len & H/E & Chk & Loop & Hdg & Len & H/E & Chk & Loop & Hdg & TS \\
\midrule
Qwen3.5-9B & 23.4 & 3.6 & 3.09 & 38 & 47 & 9.6 & 4.5 & 3.89 & 14 & 34 & 18.0 & 2.9 & 2.45 & 26 & 52 & 7.2 \\
\midrule
\tgroup{17}{Mixed training on all domains} \\
MixSFT & 3.5 & 2.6 & 1.69 & 2 & 20 & 0.9 & 1.6 & 0.93 & 0 & 10 & 2.7 & 2.2 & 1.31 & 1 & 10 & 4.1 \\
+ MixRL (MixRL baseline) & 4.8 & 3.1 & 1.98 & 3 & 29 & 1.5 & 2.7 & 3.42 & 1 & 16 & 1.4 & 3.7 & 2.00 & 0 & 8 & 0.9 \\
\midrule
\tgroup{17}{Domain teachers with the SFT warm-up} \\
SFT warm-ups$^\dagger$ & 2.8 & 2.6 & 2.02 & 1 & 19 & 1.1 & 1.5 & 0.96 & 0 & 10 & 0.8 & 1.2 & 0.37 & 0 & 2 & 2.4 \\
+ GSPO (domain teachers)$^\dagger$ & 6.4 & 3.5 & 2.11 & 2 & 35 & 1.0 & 2.8 & 2.63 & 0 & 8 & 0.7 & 1.3 & 0.64 & 0 & 2 & 2.5 \\
+ MOPD & 6.3 & 3.4 & 1.98 & 2 & 34 & 1.6 & 4.4 & 2.66 & 0 & 13 & 0.7 & 1.2 & 0.33 & 0 & 2 & 2.7 \\
+ MixRL & 6.2 & 3.8 & 2.31 & 1 & 34 & 1.1 & 4.1 & 2.92 & 0 & 10 & 1.1 & 3.2 & 2.20 & 0 & 3 & 3.4 \\
\midrule
\tgroup{17}{Domain teachers without the SFT warm-up} \\
GSPO (domain teachers)$^\dagger$ & 11.3 & 3.3 & 2.57 & 16 & 44 & 3.1 & 4.0 & 2.72 & 4 & 22 & 2.5 & 3.2 & 2.51 & 3 & 22 & 4.6 \\
+ MOPD & 11.1 & 3.1 & 2.47 & 17 & 43 & 3.5 & 4.2 & 2.64 & 4 & 25 & 2.5 & 3.6 & 2.58 & 3 & 22 & 4.5 \\
\textbf{+ MixRL (\oursmodel)} & 11.7 & 3.3 & 2.48 & 17 & 42 & 3.5 & 5.0 & 2.65 & 4 & 22 & 1.2 & 4.4 & 2.80 & 2 & 4 & 5.2 \\
\bottomrule
\end{tabular}
\end{table*}
\else\input{tables/analysis_v2_behavior_main}\fi  %
\ifdefined\ppmode\paragraph{Measures.}
We study the reasoning of the models at the level of their outputs and treat the thinking part of a response as a sequence of reasoning moves.
We derive the moves from the responses: we collect the most frequent sentence openings in the responses of all models, group them into moves, and check each group by reading samples of the matched sentences.
Similar to the cognitive behaviors of Gandhi et al.~\cite{gandhi2025cognitive}, we count three moves of self-monitoring, namely verifying a result (\eg ``let's double-check'', ``this matches''), revising a step (\eg ``wait'', ``let me reconsider''), and considering an alternative (\eg ``or maybe'', ``let's assume''), which we call self-checks.
Since our models express the same move either in the native style of Qwen3.5-9B (``Let's check'') or in the first-person style of our SFT traces (``Let me check'', ``I need to''), each move covers both phrasings, and we normalize the count by the thinking length.
We further describe the allocation of the thinking budget as the ratio of the thinking length on hard questions to that on easy questions, where we rank the questions by the mean solve rate of five models\ifdefined\ppmode\space(the domain teachers trained directly from Qwen3.5-9B, their MOPD student, \oursmodel, MixSFT and the MixRL baseline, with easy questions at a solve rate of at least 0.9 and hard questions at most 0.25)\fi, the stability as the share of responses with at least five repeated lines, the decisiveness as the share of responses that hedge (\eg ``probably'', ``I'll go with''), and, on video, the grounding as the number of cited timestamps.
We compute all measures on the logged response of each question, which is the first of the eight samples of the avg@8 evaluation (Table~\ref{tab:behavior}).

\fi\paragraph{Reasoning behavior.}
\ifdefined\ppmode We discuss four of these measures here, and Appendix~F discusses the self-checks, the loops and the other stages (Table~\ref{tab:behavior}).\else We describe the reasoning of each model at the level of its outputs with four measures, computed on one logged response per benchmark question (Table~\ref{tab:behavior_main}).\fi{}
The thinking length shows how much budget the model spends, the ratio of this length on hard questions to that on easy questions shows whether it spends the budget where it is needed, the share of responses that hedge shows how firmly it decides, and, on video, the density of cited timestamps shows how much it grounds its reasoning in the clip.
Qwen3.5-9B overthinks, and this holds on video as well, where it thinks at length, many of its responses hedge, and it cites many moments of the clip before it decides.
The MixRL baseline is the opposite case, with short answers on all domains and few cited moments.
The teachers of our recipe set a budget for each domain, long on image STEM and far shorter than that of Qwen3.5-9B on video, and the MOPD student stays close to its teachers in every measure, which shows that MOPD transfers how each teacher reasons and not only what it answers.

On top of this student, MixRL keeps the thinking length on the two image domains.
On video, in contrast, MixRL replaces the long, timestamped evidence chains of the student with short and direct answers, so the thinking becomes much shorter, hedging responses become rare, and the thinking length depends more on the difficulty of the question.
The density of cited timestamps, however, stays above the level of the student and far above that of the MixRL baseline.
On top of the MOPD student of the teachers with the SFT warm-up, whose thinking length on video hardly depends on the difficulty, MixRL makes it depend clearly on the difficulty, while the responses stay short.
These changes match the gains in Section~\ref{sec:main_results}, which are largest on video for both students.
MixRL on top of MOPD thus keeps the skills that MOPD transfers from the teachers and refines them on the domains where the teachers leave room.

\begin{table}[t]
\centering\footnotesize
\setlength{\tabcolsep}{3.5pt}
\renewcommand{\arraystretch}{0.92}
\caption{MixRL on top of MOPD with and without a prompt filter, for the teachers with the SFT warm-up (domain means and overall mean, avg@8 accuracy, \%). $h$ is the number of solved responses ($R_\text{acc}\ge 0.5$) of the MOPD student among eight samples of a prompt.}
\label{tab:prompt_filter}
\begin{tabular}{@{}lrcccc@{}}
\toprule
Prompts & Size & STEM & Perc. & Video & Overall \\
\midrule
MOPD student (start) & -- & 74.8 & 80.6 & 62.8 & 74.2 \\
\midrule
Full union & 193,027 & 75.3 & 81.2 & 63.7 & 74.8 \\
Drop $h\in\{0,8\}$ & 56,713 & 75.4 & 80.6 & 63.6 & 74.6 \\
Keep $h\le 4$ & 60,863 & 74.8 & 80.6 & 62.6 & 74.1 \\
\bottomrule
\end{tabular}
\end{table}

\paragraph{Prompt filtering.}
Because MixRL on top of MOPD uses the RL prompts a second time, we also examine whether it needs all of them.
With the MOPD student of the teachers with the SFT warm-up, we sample eight responses for each prompt of the RL training set and let $h$ be the number of solved responses.
We then compare the full prompt union with two filtered sets, one without the prompts that the student never or always solves ($h\in\{0,8\}$) and one with only the prompts that it does not yet solve reliably ($h\le 4$), where all runs start from the same MOPD checkpoint and use the same settings (Table~\ref{tab:prompt_filter}).
We find that neither filter improves on the full union.
The run without the prompts with $h\in\{0,8\}$ still improves the student and stays close to the full union with far fewer prompts, whereas the run that keeps only the prompts with $h\le 4$ does not improve the student.
For a fair comparison with the mixed-training MixRL runs, we use the full union for the main results.

\section{Conclusion}\label{sec:conclusion}

We presented \ours, a fully open recipe for reasoning VLMs over image STEM, image perception and video.
On the same curated data, MOPD of GSPO teachers trained directly from Qwen3.5-9B is the best of the compared recipes, and with or without the SFT warm-up, MOPD and MixRL complement each other, as MOPD merges the domain teachers into a stronger starting point for RL than the base model, and MixRL still improves this student on the same data with no collapse.
We \ifdefined\ppmode release\else will release\fi{} all data, the training code, and the checkpoints of every stage.
\par

{
    \small
    \bibliographystyle{ieeenat_fullname}
    \bibliography{main}
}

\providecommand{\suppstart}{\clearpage\setcounter{page}{1}\maketitlesupplementary}
\suppstart
\providecolor{oursL}{HTML}{EEF3FA}   %
\providecolor{overallL}{HTML}{E8E8E8}
\providecommand{\svAccW}{0.9}
\providecommand{\svFmtW}{0.1}
\providecommand{\svPartialBox}{0.5}
\providecommand{\svAnls}{0.5}
\providecommand{\svCoordScale}{1000}
\providecommand{\svPointRadius}{20}
\providecommand{\svMaxAnsChars}{300}
\providecommand{\svMathTimeout}{3}
\providecommand{\svNgram}{13}
\providecommand{\svPhash}{4}
\providecommand{\svEvalCap}{32768}
\providecommand{\svEvalT}{0.6}
\providecommand{\svEvalTopP}{0.95}
\providecommand{\svEvalN}{8}
\providecommand{\svEvalTopK}{20}
\providecommand{\svImgMin}{65,536}
\providecommand{\svImgMax}{1,605,632}
\providecommand{\svFps}{1}
\providecommand{\svFrMin}{4}
\providecommand{\svFrMax}{150}
\providecommand{\svFrPxMin}{4,096}
\providecommand{\svFrPxMax}{65,536}
\providecommand{\svTopK}{32}
\providecommand{\svClamp}{10}
\providecommand{\svLpMin}{$-10$}
\providecommand{\svMopdBatch}{1008}
\providecommand{\svMopdSteps}{191}
\providecommand{\svMopdPrompts}{193,027}
\providecommand{\svMopdResp}{16384}
\providecommand{\svStale}{2}
\providecommand{\svTeacherMml}{22017}
\providecommand{\svMopdPrompt}{5632}
\providecommand{\svGspoN}{8}
\providecommand{\svSftTokStep}{1,048,576}
\providecommand{\svVerlVer}{0.9.1}
\providecommand{\svVllmVer}{0.19.1}
\providecommand{\svEngineVer}{0.1.4}
\providecommand{\svEvalVer}{0.7.2}
\providecommand{\svMvVer}{0.9.0}
\providecommand{\svTorchVer}{2.10.0}
\providecommand{\svPoolSeed}{20261022}
\providecommand{\svGenSelfT}{0.7}
\providecommand{\svKeepSelf}{78.1}
\providecommand{\svKeepNext}{99.1}
\providecommand{\svMixRLStep}{240}

\appendix
\ifdefined\ppmode\else   %
\setcounter{table}{0}\renewcommand{\thetable}{S\arabic{table}}
\setcounter{figure}{0}\renewcommand{\thefigure}{S\arabic{figure}}
\setcounter{equation}{0}\renewcommand{\theequation}{S\arabic{equation}}
\fi

\setcounter{tocdepth}{1}
\tableofcontents

\section{Data}\label{sec:suppl_data}

\subsection{Sources}\label{sec:suppl_sources}
Each domain has an SFT set and an RL prompt set built from public datasets (Table~\ref{tab:supp_data}).
For the SFT sets, we keep the questions and gold answers of the sources and replace the original responses with samples from a model that we give the gold final answer.
The SFT question pool contains only single-turn questions with at least one image or video, and multi-turn sources keep their last question.
In contrast, the RL sets keep only prompts with rule-verifiable answers, and Table~\ref{tab:supp_tasktypes} gives their answer types.
Within one source, a stable hash of the image, video or question splits the items between SFT, RL and a held-out pool, and the SFT part of a source excludes the images of its RL part.
Finally, several RL sources are subsampled, and the pass-rate filters of DeepVision-103K and ViRL39K keep the harder items.

\begin{table*}[!htbp]
\centering
\footnotesize
\renewcommand{\arraystretch}{1.04}
\setlength{\tabcolsep}{3pt}
\caption{Data sources and sizes after deduplication and decontamination. ``SFT'' is the SFT set of the domain, which serves the SFT warm-up of the domain teacher that we compare and is a part of the MixSFT set, and ``RL'' the RL prompt set of the domain teacher. The union of the three RL sets is the prompt set of MOPD and MixRL. The image SFT sets carry self traces of Qwen3.5-9B, and the video SFT set carries traces of Qwen3.8-Flash-Next~\cite{qwen2026qwen38flashnext}. DeepVision-103K keeps math items with a pass rate in $(0, 0.5]$ and ViRL39K items with a pass rate of at most 0.25 (hard items). \textsuperscript{a}The RL part holds out the DocVQA, TextVQA, ST-VQA, InfographicVQA questions of FineVision (15\% of their images by hash).}
\label{tab:supp_data}
\resizebox{\textwidth}{!}{%
\begin{tabular}[t]{@{}lrr@{}}
\toprule
\textbf{Source} & \textbf{SFT} & \textbf{RL} \\
\midrule
\rowcolor{domImgrL}\multicolumn{3}{@{}l}{\textbf{Image STEM}} \\
MMFineReason-SFT~\cite{lin2026mmfinereason} & 323,510 & -- \\
MMVistaReason-SFT (math, science, logic)~\cite{lin2026mmvistareason} & 30,990 & -- \\
FineVision~\cite{wiedmann2025finevision}: TabMWP, ScienceQA, CLEVR-Math & 13,802 & -- \\
DeepVision-103K~\cite{sun2026deepvision} & -- & 23,486 \\
Vero-600K, STEM subsets~\cite{sarch2026vero} & -- & 18,093 \\
MMVistaReason-RL~\cite{lin2026mmvistareason} & -- & 17,614 \\
ViRL39K~\cite{wang2025vlrethinker} & -- & 5,114 \\
\textit{Total} & \textit{368,302} & \textit{64,307} \\
\rowcolor{domVidL}\multicolumn{3}{@{}l}{\textbf{Video}} \\
LLaVA-Video-178K~\cite{zhang2024llavavideo} & 45,800 & -- \\
LongVideo-Reason~\cite{chen2025longvideo} & 19,889 & 3,170 \\
OneThinker~\cite{feng2025onethinker} & 16,490 & 14,624 \\
Video-R2~\cite{maaz2025videor2} & 3,696 & 2,052 \\
VideoChat3-Academic~\cite{li2026videochat3} & 270,390 & -- \\
VideoKR~\cite{fu2026videokr} & 1,039 & 125 \\
Video-HopChain~\cite{quang2026videohopchain} & -- & 14,000 \\
CapRL-Video-QA~\cite{yang2026caprlpp} & -- & 9,964 \\
TimeLens2~\cite{zhu2026timelens2} & -- & 7,000 \\
PyVision-Video-RL~\cite{zhao2026pyvision} (from SpaceR~\cite{ouyang2025spacer}) & -- & 5,870 \\
TimeLens-100K~\cite{zhang2026timelens} & -- & 4,000 \\
\textit{Total} & \textit{357,304} & \textit{60,805} \\
\bottomrule
\end{tabular}
\hspace{10pt}
\begin{tabular}[t]{@{}lrr@{}}
\toprule
\textbf{Source} & \textbf{SFT} & \textbf{RL} \\
\midrule
\rowcolor{domImgpL}\multicolumn{3}{@{}l}{\textbf{Image perception}} \\
ChartNet~\cite{kondic2026chartnet} & 36,164 & -- \\
ChartVerse-SFT~\cite{liu2026chartverse} (reasoning questions) & 23,119 & -- \\
CoSyn-point~\cite{yang2025cosyn} & 31,664 & -- \\
ECD-10k~\cite{yang2025ecd} & 20,606 & -- \\
FineVision~\cite{wiedmann2025finevision}: charts and tables & 57,749 & -- \\
FineVision~\cite{wiedmann2025finevision}: documents & 60,521 & -- \\
FineVision~\cite{wiedmann2025finevision}: general VQA & 42,294 & -- \\
FineVision~\cite{wiedmann2025finevision}: grounding, counting & 50,600 & -- \\
FineVision~\cite{wiedmann2025finevision}: hallucination robustness & 14,703 & -- \\
FineVision~\cite{wiedmann2025finevision}: OCR and document VQA\textsuperscript{a} & 94,183 & 7,979 \\
LLaVA-OneVision-1.5~\cite{an2025llavaonevision} (GQA) & 12,550 & -- \\
LLaVA-OneVision-2~\cite{an2026llavaonevision2} (spatial) & 47,652 & -- \\
SAT~\cite{ray2025sat} (synthetic 3D spatial) & 12,356 & -- \\
FineVision~\cite{wiedmann2025finevision}: general instruction data & 97,330 & -- \\
FineVision~\cite{wiedmann2025finevision}: OCR-VQA, TextVQA, ST-VQA, TextCaps, ScreenQA & 24,502 & -- \\
Vero-600K, counting, grounding, search~\cite{sarch2026vero} & -- & 14,991 \\
Vero-600K, chart and OCR~\cite{sarch2026vero} & -- & 14,959 \\
ChartVerse~\cite{liu2026chartverse} (numeric questions) & -- & 12,000 \\
Vero-600K, spatial~\cite{sarch2026vero} & -- & 7,994 \\
ZwZ-RL-VQA~\cite{wei2026zooming} & -- & 6,000 \\
LLaVA-OneVision-1.5~\cite{an2025llavaonevision} RL & -- & 3,992 \\
\textit{Total} & \textit{625,993} & \textit{67,915} \\
\midrule
\textbf{All domains} & \textbf{1,351,599} & \textbf{193,027} \\
\bottomrule
\end{tabular}
}
\end{table*}

\begin{table}[!htbp]
\centering
\small
\renewcommand{\arraystretch}{1.06}
\setlength{\tabcolsep}{3.5pt}
\caption{RL prompts by answer type. The answer type selects the format instruction and the answer matcher (Table~\ref{tab:supp_reward}).}
\label{tab:supp_tasktypes}
\begin{tabular}{@{}lrrrr@{}}
\toprule
\textbf{Answer type} & \cellcolor{domImgrL}\textbf{Image STEM} & \cellcolor{domImgpL}\textbf{Image perception} & \cellcolor{domVidL}\textbf{Video} & \textbf{All} \\
\midrule
multiple choice & 28,889 & 10,831 & 30,853 & 70,573 \\
math & 30,732 & 23,643 & 2,071 & 56,446 \\
short answer & 4,522 & 6,479 & -- & 11,001 \\
counting & 164 & 7,484 & 14,000 & 21,648 \\
OCR & -- & 9,979 & -- & 9,979 \\
grounding (box/point) & -- & 9,499 & -- & 9,499 \\
temporal grounding & -- & -- & 13,881 & 13,881 \\
\midrule
Total & 64,307 & 67,915 & 60,805 & 193,027 \\
\bottomrule
\end{tabular}
\end{table}

\subsection{Prompt format}\label{sec:suppl_prompt}
All stages and the evaluation use the native Qwen3.5 chat template with thinking enabled and no system prompt.
The user turn holds the media first and then the question, followed by one format instruction for the answer type of the prompt (Table~\ref{tab:supp_reward}).
Before we add this instruction, we remove answer-format instructions that the source questions already contain, such as requests for \texttt{<think>} or \texttt{<answer>} tags, an empty box, or ``Answer with a single word''.

\subsection{Deduplication and decontamination}\label{sec:suppl_decontam}
In the SFT question pool, we remove samples with near-identical media and the same normalized question across all domains, and the SFT and RL sets are disjoint on the prompt and media hash.
We then decontaminate every set against the evaluation benchmarks with the rules of Table~\ref{tab:supp_decontam}: exact and \svNgram-gram text overlap, image hashes with a perceptual-hash distance of at most \svPhash, and video identifiers and frame hashes.
The prompt set of MOPD and MixRL is exactly the union of the three teacher RL sets, which forms \ourrl.

\begin{table}[!htbp]
\centering
\small
\renewcommand{\arraystretch}{1.06}
\setlength{\tabcolsep}{4pt}
\caption{Decontamination against the evaluation benchmarks. Questions shorter than 50 characters are not indexed for text matching, and $n$-grams shared by more than 20 training rows count as templates.}
\label{tab:supp_decontam}
\begin{tabular}{@{}lccc@{}}
\toprule
\textbf{Check} & \multicolumn{3}{c}{\textbf{Rule}} \\
\midrule
Text, exact & \multicolumn{3}{l}{same normalized question} \\
Text, $n$-gram & \multicolumn{3}{l}{13-gram overlap $\ge$ 0.5 of the item (or $\ge$ 0.9 incl.\ templates)} \\
Image, hash & \multicolumn{3}{l}{same SHA-1} \\
Image, pHash & \multicolumn{3}{l}{64-bit distance $\le$ 4, confirmed by 256-bit $\le$ 10} \\
Video, id & \multicolumn{3}{l}{same video id (YouTube id of clips)} \\
Video, frames & \multicolumn{3}{l}{pHash of frames at 0.25, 0.5, 0.75 of the length, $\le$ 4} \\
\midrule
\textbf{Dropped rows} & \cellcolor{domImgrL}\textbf{Image STEM} & \cellcolor{domImgpL}\textbf{Image perception} & \ifdefined\ppmode\multicolumn{1}{>{\columncolor{domVidL}[\tabcolsep][0pt]}c@{}}{\textbf{Video}}\else\cellcolor{domVidL}\textbf{Video}\fi \\
\midrule
SFT & 45 & 158 & 0 \\
RL & 355 & 77 & 592 \\
\bottomrule
\end{tabular}
\end{table}

\subsection{Trace source of the SFT data}\label{sec:suppl_trace}

\ifdefined\ppmode\else\begin{table}[t]
\centering\footnotesize
\setlength{\tabcolsep}{4pt}
\renewcommand{\arraystretch}{1.05}
\caption{SFT on Qwen3.8-Flash-Next traces and on self traces of Qwen3.5-9B (domain mean of the SFT model of each domain on the benchmarks of its domain, avg@8 accuracy, \%). $\Delta$ is the change from the left to the right value.}
\label{tab:ablation_teacher}
\begin{tabular}{@{}lccc@{}}
\toprule
\textcolor{tabSub}{\footnotesize Domain (avg@8, \%)} & \textcolor{tabSub}{\footnotesize Qwen3.8-Flash-Next traces} & \textcolor{tabSub}{\footnotesize Self traces} & \textcolor{tabSub}{\footnotesize $\Delta$} \\
\midrule
\domdot{domImgr}\,Image STEM & 60.6 & 63.4 & \dchip{p}{2.8} \\
\domdot{domImgp}\,Image perception & 77.2 & 77.7 & \dchip{p}{0.5} \\
\domdot{domVid}\,Video & 59.2 & 59.1 & \dchip{n}{0.1} \\
\bottomrule
\end{tabular}
\end{table}
\fi

\ifdefined\ppmode\else
For each domain, we generate the SFT traces with two generators, Qwen3.5-9B itself (self traces) and the larger Qwen3.8-Flash-Next, and we train one SFT model of the domain on each trace source.
Table~\ref{tab:ablation_teacher} compares the two models on the benchmarks of their domain.
On both image domains, self traces give the better model, with a clear margin on image STEM, whereas on video, self traces give a slightly worse model than the Qwen3.8-Flash-Next traces.
This agrees with the observation from OPD that a stronger teacher is not always a better one~\cite{acikgoz2026synergy,li2026rethinkingopd,ma2026mopd}: on images, the traces of the model itself stay close to its own response distribution and serve it better than the traces of a stronger model.
We therefore use self traces for the two image domains and Qwen3.8-Flash-Next traces for video.
\fi
The trace filters (Section~3.1\ofmain{}) keep \svKeepSelf\% of the image self traces and \svKeepNext\% of the Qwen3.8-Flash-Next video traces.
In addition, MixSFT stays close to the SFT warm-ups on each domain (Table~\mref{tab:stages}{1}\ofmain{}), so the MixRL baseline and the teachers with the SFT warm-up start from SFT checkpoints of similar quality.

The trace source also shapes the reasoning style that the later RL keeps (Section~\ref{sec:suppl_behavior}).
We write the SFT traces with the gold answer given, so they are short and contain few self-checks, and the SFT models take over this profile.
RL from these models adds some self-checks on all domains and makes the responses longer on image STEM, but the responses remain much shorter than those of the teachers trained directly from Qwen3.5-9B.

\section{Reward and Answer Matching}\label{sec:suppl_reward}

\ifdefined\ppmode Section~\ref{sec:recipe_reward} defines the reward, and this section gives the answer matchers.\space\else
All RL stages use one rule-based reward, $R = \svAccW\,R_\text{acc} + \svFmtW\,R_\text{fmt}$, where $R_\text{acc}\in[0,1]$ is the accuracy term and $R_\text{fmt}$ is the format term.
The format term is 1 for a response with a valid thinking block and exactly one box in its answer part, \svPartialBox{} for a response with a valid thinking block but not exactly one box, and 0 otherwise.
A valid thinking block means that the response closes the block exactly once, with non-empty reasoning before \texttt{</think>} and a non-empty answer after it, and the format term scores the whole response once, not each token.
In addition, the accuracy term is set to zero unless the format term is 1.
\fi
Before matching, we normalize the prediction and the gold answer by removing math delimiters, bold markers, text macros, quotes, a leading ``The answer is'' and a trailing period, and by writing percent and degree signs in one form.

Table~\ref{tab:supp_reward_types} lists the answer types of the RL prompt sets with the rule that scores each type.
The answer matcher depends on the answer type of the prompt (Table~\ref{tab:supp_reward}).
For math, we use the Math-Verify library with numeric fallbacks: units must agree when both sides give one, percentages match decimals, multi-part answers match part by part, and a stated relation (``$\geq 5$'') never matches a plain value.
For short answers, we match the normalized strings, and a numeric short answer follows the rule for math.
For multiple choice, we accept option letters in common forms and sets of letters for multi-answer questions.
For counting, we accept number words and approximate phrases (``about 7'').
For OCR, we use ANLS with a threshold of \svAnls, and different digit sequences score zero.
For grounding, boxes use coordinates in $[0,\svCoordScale]$, and the score is the Hungarian-matched IoU sum divided by the larger of the predicted and gold box counts.
A point gold matches when the predicted point is within \svPointRadius{} units or inside the predicted box.
For temporal grounding, we use the matched temporal IoU over the predicted segments.
Finally, before Math-Verify, the prediction is capped at \svMaxAnsChars{} characters (more for long gold answers), and a Math-Verify call times out after \svMathTimeout{} seconds.

\begin{table}[!htbp]
\centering
\small
\renewcommand{\arraystretch}{1.18}
\setlength{\tabcolsep}{3pt}
\caption{Answer types of the RL prompt sets and the rule that scores each type. The accuracy term $R_\text{acc}$ is binary for the first four types and continuous in $[0,1]$ for the last three.}
\label{tab:supp_reward_types}
\begin{tabular}{@{}llrrr@{}}
\toprule
\textbf{Type} & \textbf{Matching rule} & \cellcolor{domImgrL}\textbf{Image STEM} & \cellcolor{domImgpL}\textbf{Image perc.} & \ifdefined\ppmode\multicolumn{1}{>{\columncolor{domVidL}[\tabcolsep][0pt]}r@{}}{\textbf{Video}}\else\cellcolor{domVidL}\textbf{Video}\fi \\
\midrule
Math & Math-Verify, units, \% = decimal & 30.7K & 23.6K & 2.1K \\
Mult.\ choice & option letter or letter set & 28.9K & 10.8K & 30.9K \\
Short & normalized text or number & 4.5K & 6.5K & -- \\
Count & integer or number word & 164 & 7.5K & 14.0K \\
OCR & ANLS $\ge$ \svAnls, digit guard & -- & 10.0K & -- \\
Box/point & matched IoU, point radius & -- & 9.5K & -- \\
Segment & matched temporal IoU & -- & -- & 13.9K \\
\midrule
Total & & 64.3K & 67.9K & 60.8K \\
\bottomrule
\end{tabular}
\end{table}

\begin{table*}[!htbp]
\centering
\small
\renewcommand{\arraystretch}{1.06}
\setlength{\tabcolsep}{6pt}
\caption{Answer types, format instructions and answer matchers of the rule-based reward (Math-Verify 0.9.0). The reward is $R = 0.9\,R_\text{acc} + 0.1\,R_\text{fmt}$, and $R_\text{acc}$ counts only when $R_\text{fmt} = 1$.}
\label{tab:supp_reward}
\begin{tabular}{@{}lp{5.5cm}p{2.9cm}p{5.0cm}@{}}
\toprule
\textbf{Answer type} & \textbf{Format instruction (appended to the question)} & \textbf{Matcher} & \textbf{Details} \\
\midrule
multiple choice & \texttt{\small Answer with the option letter only, inside \textbackslash boxed\{\}.} & option letter or set of letters & letter forms (``B'', ``(B)'', ``B. text''); multi-answer sets \\
math & \texttt{\small Put your final answer within \textbackslash boxed\{\}.} & Math-Verify + numeric fallback & units must agree, \% $\leftrightarrow$ decimal, relations never match a value; rounding 6 digits \\
short answer & \texttt{\small Put your final short answer within \textbackslash boxed\{\}.} & normalized string + numeric fallback & as math for numbers; multi-part answers part by part \\
counting & \texttt{\small Put your final answer as a single integer within \textbackslash boxed\{\}.} & integer & number words, ``about 7'', ``7 birds'' \\
OCR & \texttt{\small Put the exact text of your final answer within \textbackslash boxed\{\}.} & ANLS & threshold 0.5; different digit sequences score 0 \\
grounding (box/point) & \texttt{\small Output the box as [x1, y1, x2, y2] in 0-1000 relative coordinates inside \textbackslash boxed\{\}.} & Hungarian-matched IoU & coordinates in $[0, 1000]$; sum IoU / max(\#pred, \#gold); point gold: within 20 units or inside the box \\
temporal grounding & \texttt{\small Output the segment as [start, end] in seconds inside \textbackslash boxed\{\}.} & matched temporal IoU & seconds or mm:ss; several segments matched, / max(\#pred, \#gold) \\
\bottomrule
\end{tabular}
\end{table*}

\section{Evaluation Protocol}\label{sec:suppl_eval}

We evaluate all models in lmms-eval~\cite{zhang2024lmmseval} (version \svEvalVer) with the vLLM backend~\cite{kwon2023vllm} and one protocol.
We sample \svEvalN{} responses per question (temperature \svEvalT, top-$p$ \svEvalTopP, top-$k$ \svEvalTopK) and report the mean accuracy over them (avg@\svEvalN), with the native Qwen3.5 chat template, thinking enabled and no system prompt.
The response cap is \svEvalCap{} tokens, thinking included.
Images have between \svImgMin{} and \svImgMax{} pixels.
Videos shorter than \svFrMax{} seconds are sampled at \svFps{} fps with at least \svFrMin{} frames, and from longer videos we sample \svFrMax{} evenly spaced frames.
Each frame has \svFrPxMin{} to \svFrPxMax{} pixels, the frames are passed to the model as one video, and each frame carries the timestamp text of the Qwen3.5 processor.

Domain means average all benchmarks of a domain, and the overall mean averages all 21 benchmarks.
We-Math reports its strict score, and on SimpleVQA, a response counts as correct when it matches the gold answer after normalization or when a judge model, Qwen3.8-27B~\cite{qwen2026qwen38}, grades it as correct.
Table~\ref{tab:supp_eval} lists the dataset, split and size of every benchmark.

\begin{table*}[!htbp]
\centering
\footnotesize
\renewcommand{\arraystretch}{1.06}
\setlength{\tabcolsep}{5pt}
\caption{Evaluation benchmarks. $n$ is the number of scored items. DynaMath reports the accuracy over all generated variants.}
\label{tab:supp_eval}
\begin{tabular}{@{}llllr@{}}
\toprule
\textbf{Benchmark} & \textbf{Dataset} & \textbf{Subset} & \textbf{Split} & \textbf{$n$} \\
\midrule
\rowcolor{domImgrL}\multicolumn{5}{@{}l}{\textbf{Image STEM}} \\
DynaMath~\cite{zou2025dynamath} & \texttt{\footnotesize kcz358/DynaMath} & -- & test & 5,010 \\
EMMA~\cite{hao2025emma} & \texttt{\footnotesize lmms-lab/EMMA} & All & test & 2,788 \\
LogicVista~\cite{xiao2024logicvista} & \texttt{\footnotesize lscpku/LogicVista} & -- & test & 448 \\
MathVerse~\cite{zhang2024mathverse} & \texttt{\footnotesize CaraJ/MathVerse-lmmseval} & testmini & testmini & 3,940 \\
MathVista~\cite{lu2024mathvista} & \texttt{\footnotesize AI4Math/MathVista} & -- & testmini & 1,000 \\
MMMU~\cite{yue2024mmmu} & \texttt{\footnotesize lmms-lab-encoder/MMMU} & -- & validation & 900 \\
MMMU-Pro~\cite{yue2025mmmupro} & \texttt{\footnotesize MMMU/MMMU\_Pro} & standard (10 options) & test & 1,730 \\
We-Math~\cite{qiao2024wemath} & \texttt{\footnotesize We-Math/We-Math} & -- & testmini & 1,740 \\
\rowcolor{domImgpL}\multicolumn{5}{@{}l}{\textbf{Image perception}} \\
AI2D~\cite{kembhavi2016ai2d} & \texttt{\footnotesize lmms-lab-encoder/ai2d} & -- & test & 3,088 \\
CountBenchQA~\cite{paiss2023countbench,beyer2024paligemma} & \texttt{\footnotesize vikhyatk/CountBenchQA} & -- & test & 491 \\
CV-Bench~\cite{tong2024cambrian} & \texttt{\footnotesize nyu-visionx/CV-Bench} & default & test & 2,638 \\
HallusionBench~\cite{guan2024hallusionbench} & \texttt{\footnotesize lmms-lab-encoder/HallusionBench} & -- & image & 951 \\
MMStar~\cite{chen2024mmstar} & \texttt{\footnotesize Lin-Chen/MMStar} & -- & val & 1,500 \\
OCRBench~\cite{liu2024ocrbench} & \texttt{\footnotesize echo840/OCRBench} & -- & test & 1,000 \\
RealWorldQA~\cite{xai2024realworldqa} & \texttt{\footnotesize lmms-lab-encoder/RealWorldQA} & -- & test & 765 \\
SimpleVQA~\cite{cheng2025simplevqa} & \texttt{\footnotesize m-a-p/SimpleVQA} & -- & test & 2,025 \\
\rowcolor{domVidL}\multicolumn{5}{@{}l}{\textbf{Video}} \\
LongVideoBench~\cite{wu2024longvideobench} & \texttt{\footnotesize longvideobench/LongVideoBench} & -- & validation & 1,337 \\
TempCompass~\cite{liu2024tempcompass} & \texttt{\footnotesize lmms-eval/TempCompass} & multi-choice & test & 1,580 \\
Video-Holmes~\cite{cheng2025videoholmes} & \texttt{\footnotesize TencentARC/Video-Holmes} & -- & test & 1,837 \\
Video-MME~\cite{fu2025videomme} & \texttt{\footnotesize lmms-eval/Video-MME} & -- & test & 2,700 \\
VSI-Bench~\cite{yang2025thinkinginspace} & \texttt{\footnotesize nyu-visionx/VSI-Bench} & full & test & 5,130 \\
\bottomrule
\end{tabular}
\end{table*}

\section{Training Configurations}\label{sec:suppl_config}

Table~\ref{tab:hparams} gives an overview of the settings of all stages, and Tables~\ref{tab:supp_sft}, \ref{tab:supp_rl} and~\ref{tab:supp_mopd} give the full configurations of SFT, GSPO and MOPD.
We train SFT with lmms-engine~\cite{lmmsengine2025} (version \svEngineVer), and GSPO and MOPD with verl~\cite{sheng2025hybridflow} (version \svVerlVer) and vLLM~\cite{kwon2023vllm} (version \svVllmVer) for rollout, all on PyTorch \svTorchVer.
All experiments run on a pool of 64 80GB GPUs, and the released configurations launch every stage on four nodes of eight GPUs, with two training and two rollout nodes for GSPO and two training, one rollout and one teacher node for MOPD.
We train each run until its training metric converges.
All models build on Qwen3.5-9B~\cite{qwen2026qwen35}, use full-parameter training, and share the media settings of the evaluation protocol.
SFT uses about \svSftTokStep{} tokens per optimizer step in every run, with packed sequences.
The GSPO runs of the three teachers and the MixRL baseline share all objective, rollout and optimization settings.
Table~\ref{tab:supp_rl} shows the teachers trained directly from Qwen3.5-9B. The teachers with the SFT warm-up use the same objective, rollout and optimization settings and start from the SFT warm-ups of Table~\ref{tab:supp_sft}, and MOPD uses the settings of Table~\ref{tab:supp_mopd} for both sets of teachers.
MixRL on top of MOPD uses the objective, rollout and optimization settings of the MixRL baseline, starts from the final MOPD checkpoint, and trains on the same prompt union of \mxFullN{} prompts.
MixRL directly from Qwen3.5-9B uses the same settings and the same prompt union and starts from Qwen3.5-9B.

\begin{table}[!htbp]
\centering
\footnotesize
\setlength{\tabcolsep}{5pt}
\caption{Overview of the settings of all stages. \textsuperscript{a}MixRL on top of MOPD. The MixRL baseline starts from MixSFT and trains on the same union (Table~\ref{tab:supp_rl}). \textsuperscript{b}Packed to 8192 tokens (domain SFT) or 16384 tokens (MixSFT). \textsuperscript{c}Only for the teachers with the SFT warm-up, where GSPO starts from the SFT warm-up of the domain. Off-policy: maximum staleness in policy versions and the truncation of the token importance weight.}
\label{tab:hparams}
\begin{tabular}{@{}lcccc@{}}
\toprule
 & \multicolumn{2}{c}{Domain teachers} & & \\
\cmidrule(lr){2-3}
 & SFT warm-up\textsuperscript{c} & GSPO & MOPD & MixRL\textsuperscript{a} \\
\midrule
Init & Qwen3.5-9B & Qwen3.5-9B\textsuperscript{c} & Qwen3.5-9B & final MOPD \\
Data & domain SFT set & domain RL set & union (193K) & union (193K) \\
Learning rate & 1e-5, cosine & 1e-6, const. & 1e-6, const. & 1e-6, const. \\
Warmup & 3\% & 20 steps & 20 steps & 20 steps \\
Weight decay / grad.\ norm & 0.0/1.0 & 0.1/1.0 & 0.1/1.0 & 0.1/1.0 \\
Prompts / mini-batch & packed\textsuperscript{b} & 256/64 & 1008/1008 & 256/64 \\
Rollouts per prompt & -- & 8 & 1 & 8 \\
Temperature / top-$p$ & -- & 1.0/1.0 & 1.0/1.0 & 1.0/1.0 \\
Max.\ response & -- & 16384 & 16384 & 16384 \\
Loss & CE (assistant) & GSPO & top-32 reverse KL & GSPO \\
Clip $\epsilon_\text{l}/\epsilon_\text{h}$ / clamp & -- & 3e-4/4e-4 & $c{=}10$, floor $-10$ & 3e-4/4e-4 \\
Reward & -- & Section~\ref{sec:suppl_reward} & none & Section~\ref{sec:suppl_reward} \\
Group filter & -- & equal $R_\text{acc}$ & -- & equal $R_\text{acc}$ \\
Truncated responses & -- & scored & masked & scored \\
Reference KL & -- & none & none & none \\
Off-policy & -- & $\le$2 stale, IS 2 & $\le$2 stale & $\le$2 stale, IS 2 \\
\bottomrule
\end{tabular}
\end{table}

\begin{table*}[!htbp]
\centering
\footnotesize
\renewcommand{\arraystretch}{1.06}
\setlength{\tabcolsep}{7pt}
\caption{SFT configuration (lmms-engine 0.1.4). The three domain SFT runs are the warm-ups of the domain teachers with the SFT warm-up. MixSFT trains on the union of the three domain SFT sets and is the start of the MixRL baseline. A single centered value applies to all runs. Prompt tokens are masked from the loss on every assistant turn.}
\label{tab:supp_sft}
\begin{tabular}{@{}lcccc}
\toprule
\textbf{Setting} & \cellcolor{domImgrL}\textbf{Image STEM} & \cellcolor{domImgpL}\textbf{Image perception} & \cellcolor{domVidL}\textbf{Video} & \cellcolor{overallL}\textbf{MixSFT} \\
\midrule
\rowcolor{overallL}\multicolumn{5}{@{}l}{\textit{Data}} \\
Initialization & \multicolumn{4}{c}{Qwen3.5-9B} \\
SFT samples & 368,302 & 625,993 & 357,304 & 1,351,599 \\
Tokens (total / target) & 396.8M / 284.2M & 686.8M / 180.8M & 539.3M / 73.3M & -- \\
Trace generator & Qwen3.5-9B (self) & Qwen3.5-9B (self) & Qwen3.8-Flash-Next & as the domain runs \\
Chat template / thinking & \multicolumn{4}{c}{Qwen3.5 native / on} \\
System prompt & \multicolumn{4}{c}{none} \\
Loss & \multicolumn{4}{c}{CE on assistant tokens} \\
\rowcolor{overallL}\multicolumn{5}{@{}l}{\textit{Schedule}} \\
Packed sequence length & 8,192 & 8,192 & 8,192 & 16,384 \\
Packing strategy & \multicolumn{4}{c}{balanced} \\
Tokens per optimizer step & \multicolumn{4}{c}{1,048,576} \\
\rowcolor{overallL}\multicolumn{5}{@{}l}{\textit{Optimization}} \\
Optimizer & \multicolumn{4}{c}{AdamW (fused)} \\
Learning rate / schedule & \multicolumn{4}{c}{1e-5 / cosine} \\
Warmup ratio & \multicolumn{4}{c}{0.03} \\
Weight decay & \multicolumn{4}{c}{0.0} \\
Max.\ grad norm & \multicolumn{4}{c}{1.0} \\
Precision & \multicolumn{4}{c}{BF16, TF32, reduce BF16} \\
Sharding & \multicolumn{4}{c}{FSDP2, no reshard after fwd} \\
Grad.\ checkpointing / Liger & \multicolumn{4}{c}{on / on} \\
Attention / padding removal & \multicolumn{4}{c}{FlashAttention-2 / on} \\
Sequence parallel (Ulysses) & \multicolumn{4}{c}{none} \\
Seed (train / data) & \multicolumn{4}{c}{42 / 42} \\
\rowcolor{overallL}\multicolumn{5}{@{}l}{\textit{Media}} \\
Image pixels (min--max) & \multicolumn{4}{c}{65,536--1,605,632} \\
Video fps / frames / px per frame & \multicolumn{4}{c}{1 / 4--150 / 4,096--65,536} \\
\bottomrule
\end{tabular}
\end{table*}

\begin{table*}[!htbp]
\centering
\small
\renewcommand{\arraystretch}{1.06}
\setlength{\tabcolsep}{4.4pt}
\caption{GSPO configuration of the domain teachers trained directly from Qwen3.5-9B, the MixRL baseline, and MixRL on top of the MOPD student of these teachers (verl 0.9.1, rollout with vLLM 0.19.1). A single centered value applies to all runs. All runs use the reward of Section~\ref{sec:suppl_reward} and the media settings of the evaluation protocol (Section~\ref{sec:suppl_eval}). A step fills one batch with prompts that pass the group filter, each with eight rollouts, and makes one optimizer update per mini-batch. For both sets of teachers, MixRL on top of MOPD starts from the final checkpoint of their MOPD student and trains on the same prompts as the MixRL baseline.}
\label{tab:supp_rl}
\begin{tabular}{@{}lccccc}
\toprule
\textbf{Setting} & \cellcolor{domImgrL}\textbf{\shortstack{Image STEM\\teacher}} & \cellcolor{domImgpL}\textbf{\shortstack{Image perception\\teacher}} & \cellcolor{domVidL}\textbf{\shortstack{Video\\teacher}} & \cellcolor{overallL}\textbf{\shortstack{MixRL\\(baseline)}} & \cellcolor{oursL}\textbf{\shortstack{MixRL on top\\of MOPD}} \\
\midrule
\rowcolor{overallL}\multicolumn{6}{@{}l}{\textit{Per run}} \\
Initialization & Qwen3.5-9B & Qwen3.5-9B & Qwen3.5-9B & MixSFT & final MOPD \\
RL prompts & 64,307 & 67,915 & 60,805 & 193,027 & 193,027 \\
Prompts per step / mini-batch & 256 / 64 & 256 / 64 & 256 / 64 & 252 / 63 & 256 / 64 \\
Optimizer updates per step & 4 & 4 & 4 & 4 & 4 \\
Max.\ prompt tokens & 2,560 & 2,048 & 5,632 & 5,632 & 5,632 \\
\rowcolor{overallL}\multicolumn{6}{@{}l}{\textit{Objective}} \\
Policy loss & \multicolumn{5}{c}{GSPO} \\
Loss aggregation & \multicolumn{5}{c}{token mean per sequence, mean over sequences} \\
Clip ratio $\epsilon_\text{low}$ / $\epsilon_\text{high}$ & \multicolumn{5}{c}{3e-4 / 4e-4} \\
Advantage & \multicolumn{5}{c}{group mean, std-normalized} \\
KL loss / KL in reward & \multicolumn{5}{c}{off / off} \\
Entropy coefficient & \multicolumn{5}{c}{0} \\
Group filter (groups with equal $R_\text{acc}$) & \multicolumn{5}{c}{drop, on accuracy} \\
\rowcolor{overallL}\multicolumn{6}{@{}l}{\textit{Rollout}} \\
Rollouts per prompt & \multicolumn{5}{c}{8} \\
Temperature / top-$p$ / top-$k$ & \multicolumn{5}{c}{1 / 1 / off} \\
Max.\ response tokens & \multicolumn{5}{c}{16,384} \\
Rollout engine & \multicolumn{5}{c}{vLLM 0.19.1} \\
CUDA graphs (decode) & \multicolumn{5}{c}{on} \\
Chunked prefill / prefix cache & \multicolumn{5}{c}{on / on} \\
\rowcolor{overallL}\multicolumn{6}{@{}l}{\textit{Asynchrony}} \\
Trainer mode & \multicolumn{5}{c}{separate trainer and rollout pools, fully async} \\
Mini-batches per weight sync & \multicolumn{5}{c}{4} \\
Max.\ staleness (policy versions) & \multicolumn{5}{c}{2, older samples dropped} \\
Rollout IS correction & \multicolumn{5}{c}{token-level, truncated at 2.0} \\
Prefetched batches & \multicolumn{5}{c}{2} \\
\rowcolor{overallL}\multicolumn{6}{@{}l}{\textit{Optimization}} \\
Optimizer & \multicolumn{5}{c}{AdamW, $\beta$=(0.9, 0.999)} \\
Learning rate / schedule & \multicolumn{5}{c}{1e-6 / constant} \\
Warmup steps & \multicolumn{5}{c}{20} \\
Weight decay & \multicolumn{5}{c}{0.1} \\
Grad.\ clip (norm) & \multicolumn{5}{c}{1} \\
PPO epochs per batch & \multicolumn{5}{c}{1} \\
Precision & \multicolumn{5}{c}{FP32 master weights, BF16 compute} \\
Sharding / optimizer offload & \multicolumn{5}{c}{FSDP2, full shard / on} \\
Grad.\ checkpointing / padding removal & \multicolumn{5}{c}{on / on} \\
Dynamic batch (tokens per GPU) & \multicolumn{5}{c}{on (32,768)} \\
Sequence parallel & \multicolumn{5}{c}{none} \\
\bottomrule
\end{tabular}
\end{table*}

\ifdefined\ppmode\else\clearpage\fi  %
\section{MOPD Implementation Details}\label{sec:suppl_mopd}

Table~\ref{tab:supp_mopd} gives the configuration of MOPD, and Algorithm~\ref{alg:supp_mopd} presents one training step.

\begin{table}[!htbp]
\centering
\small
\renewcommand{\arraystretch}{1.0}
\setlength{\tabcolsep}{4pt}
\footnotesize
\caption{MOPD configuration (verl 0.9.1 with our distillation patches, vLLM 0.19.1 for student rollout and teacher scoring). The student rollout uses the sampling settings and the response limit of the GSPO runs of Table~\ref{tab:supp_rl}, with one rollout per prompt.}
\label{tab:supp_mopd}
\begin{tabular}{@{}>{\raggedright\arraybackslash}p{6.4cm}>{\raggedright\arraybackslash}p{8.6cm}}
\toprule
\textbf{Setting} & \cellcolor{oursL}\textbf{MOPD} \\
\midrule
\rowcolor{overallL}\multicolumn{2}{@{}l}{\textit{Models and data}} \\
Student initialization & Qwen3.5-9B \\
Teachers & image STEM, image perception, video GSPO checkpoints \\
Routing & hard, by the domain label of the prompt \\
Prompts & 193,027 (union of the three RL sets) \\
\rowcolor{overallL}\multicolumn{2}{@{}l}{\textit{Objective}} \\
Loss & top-$k$ reverse KL with tail correction (Eq.~\ifdefined\ppmode\ref{eq:mopd_token}\else\ref{eq:supp_tok}\fi) \\
Teacher top-$k$ & 32 \\
Per-token clamp $c$ & 10 \\
Log-prob floor (student / teacher) & $-10$ / $-10$ \\
Gradient & direct (supervised) \\
Task reward / reference KL & off / off \\
Truncated responses & masked from loss and normalizer \\
Loss aggregation & mean over all tokens of the batch \\
Rollout IS correction & not used (supervised loss); configured token-level, bounds [0.5, 5.0] \\
\rowcolor{overallL}\multicolumn{2}{@{}l}{\textit{Student rollout}} \\
Rollouts per prompt & 1 \\
Temperature / top-$p$ / top-$k$ & 1 / 1 / off \\
Max.\ prompt / response tokens & 5,632 / 16,384 \\
Prompts per step / mini-batch & 1,008 / 1,008 \\
CUDA graphs (decode) & on \\
\rowcolor{overallL}\multicolumn{2}{@{}l}{\textit{Teacher scoring}} \\
Scoring & one prefill per finished sequence, top-$k$ prompt log-probs \\
Teacher temperature & 1 \\
Teacher engine & vLLM, TP1, eager, 4,096 tokens per batch \\
Teacher context length & 22,017 \\
\rowcolor{overallL}\multicolumn{2}{@{}l}{\textit{Asynchrony and optimization}} \\
Weight sync & after every optimizer step \\
Max.\ staleness (policy versions) & 2, waits, no sample dropped \\
Prefetched batches & 1 \\
Optimizer & AdamW, $\beta$=(0.9, 0.999) \\
Learning rate / schedule / warmup & 1e-6 / constant / 20 steps \\
Weight decay / grad.\ clip & 0.1 / 1 \\
Precision & FP32 master weights, BF16 compute \\
Dynamic batch (tokens per GPU) & 22,528 \\
\bottomrule
\end{tabular}
\end{table}

\ifdefined\ppmode\else  %
\makeatletter
\@ifpackageloaded{algpseudocode}{\begin{algorithm}[!t]
\caption{One MOPD training step.}
\label{alg:supp_mopd}
\small
\begin{algorithmic}[1]
\Require student $\pi_\theta$, frozen teachers $\{\pi_d\}_{d\in\{\text{stem},\text{perc},\text{vid}\}}$, prompt union $\mathcal{D}$ with domain labels $r(x)$
\State Draw a batch $\mathcal{B}$ of 1008 prompts from $\mathcal{D}$
\ForAll{$x \in \mathcal{B}$} \Comment{student rollout, overlaps the update of the previous step}
\State Sample $y \sim \pi_{\theta'}(\cdot\mid x)$ at temperature 1, at most 16384 tokens \Comment{$\theta'$: weights at most 2 versions old}
\State Once $y$ ends, send $(x, y)$ to a replica of teacher $\pi_{r(x)}$
\State Teacher prefill: for each position $t$, the top-32 set $\mathcal{K}_t$ and $\log q(v)$, $v\in\mathcal{K}_t$, floored at $-10$
\State $m(y) \gets 0$ if $y$ hit the length cap without an end token, else $1$
\EndFor
\ForAll{$(x,y)\in\mathcal{B}$, positions $t$} \Comment{trainer}
\State $p_\theta(v) \gets \pi_\theta(v\mid x, y_{<t})$ on $\mathcal{K}_t$ (chunked over positions), $\log p_\theta(v)$ floored at $-10$ with the gradient of the unfloored value
\State $\ell_t \gets \sum_{v\in\mathcal{K}_t} \big[p_\theta(v)\log\frac{p_\theta(v)}{q(v)} - p_\theta(v) + q(v)\big]$
\State $\ell_t \gets \min(\ell_t, 10)$
\EndFor
\State $\mathcal{L} \gets \sum m(y)\sum_t \ell_t \,/\, \sum m(y)\,|y|$ \Comment{token mean over unmasked responses}
\State One AdamW update of $\theta$ on $\nabla_\theta\mathcal{L}$; sync the new weights to the rollout engines
\end{algorithmic}
\end{algorithm}
}{%
  \noindent\textit{Algorithm~\ref*{alg:supp_mopd} is typeset only in the standalone supplement.}}
\makeatother
\fi

\paragraph{Routing and teacher scoring.}
Each prompt keeps the domain label of its RL set, and the label routes the full student sequence to the frozen teacher of that domain.
The teachers share the tokenizer and processor of the student, so a teacher scores the student tokens directly, with the same media inputs.
Once a student response ends, a replica of the routed teacher scores the prompt and the response in one prefill pass and returns, for each response position $t$, the log-probabilities of its top-$k$ tokens $\mathcal{K}_t$ ($k=\svTopK$).
The teachers never decode, and scoring streams beside generation, so most of the teacher cost overlaps with the student rollout.

\paragraph{Loss.}
\ifdefined\ppmode Equation~\ref{eq:mopd_token} in Section~\ref{sec:recipe_mopd} gives the token loss.
\else
The token loss at response position $t$ is
\begin{equation}
  \ell_t(\theta) = \sum_{v\in\mathcal{K}_t} \Big[ p_\theta(v)\log\frac{p_\theta(v)}{q(v)} - p_\theta(v) + q(v) \Big],
  \label{eq:supp_tok}
\end{equation}
where $v$ is a token of the teacher top-$k$ set $\mathcal{K}_t$, $y_{<t}$ denotes the response tokens before $t$, $p_\theta(v)=\pi_\theta(v\mid x,y_{<t})$ is the student probability, and $q(v)$ is the probability of the routed teacher after the same prefix.
The tail term $-p_\theta(v)+q(v)$ makes each summand non-negative and zero only at $p_\theta(v)=q(v)$, because neither distribution sums to one on $\mathcal{K}_t$.
\fi
We floor both log-probabilities at \svLpMin{} before the loss, and the student floor passes the gradient of the unfloored value.
We compute $p_\theta$ on $\mathcal{K}_t$ in chunks of positions, so the trainer never holds the full logit tensor over the vocabulary in FP32 for a long response.
The gradient flows directly through $p_\theta$ (no policy-gradient estimator), with no task reward and no reference-model KL term.

\ifdefined\ppmode\else\paragraph{Truncated responses.}The batch loss is a token mean,
\begin{equation}
  \mathcal{L}(\theta) = \frac{\sum_{(x,y)\in\mathcal{B}} m(y) \sum_{t} \min\big(\ell_t(\theta),\, c\big)}{\sum_{(x,y)\in\mathcal{B}} m(y)\,|y|},
  \label{eq:supp_batch}
\end{equation}
where $\mathcal{B}$ is the batch of prompts $x$ with their student responses $y$, $t$ runs over the positions of $y$, $|y|$ is the token count of $y$, $c=\svClamp$ is the per-token clamp, which gives zero gradient above it, and the mask $m(y)$ is 0 if $y$ reaches the length cap of \svMopdResp{} tokens without an end token and 1 otherwise.
Without the mask, the teacher signal on a cut-off text would push the student toward text that no finished answer contains.
\fi

\paragraph{Asynchrony and steps.}
The trainer and the student rollout run on separate GPU pools.
The rollout generates the next batch while the trainer updates on the current one, and the weights sync to the rollout engines after every optimizer step.
A sample is at most \svStale{} policy versions old, and the trainer waits for in-flight samples instead of dropping them, so no student sample is discarded.
Each step takes \svMopdBatch{} prompts with one response each and makes one optimizer update.

\section{Reasoning Behavior Analysis}\label{sec:suppl_behavior}

\ifdefined\ppmode This section uses the measures of Section~5 (Table~\ref{tab:behavior}), and it analyzes\else This section gives the measures and the full numbers of the reasoning behavior in Section~5\ofmain{}, and it analyzes\fi{} the other stages of the recipe with the same measures.

\ifdefined\ppmode\else\paragraph{Measures.}
We study the reasoning of the models at the level of their outputs and treat the thinking part of a response as a sequence of reasoning moves.
We derive the moves from the responses: we collect the most frequent sentence openings in the responses of all models, group them into moves, and check each group by reading samples of the matched sentences.
Similar to the cognitive behaviors of Gandhi et al.~\cite{gandhi2025cognitive}, we count three moves of self-monitoring, namely verifying a result (\eg ``let's double-check'', ``this matches''), revising a step (\eg ``wait'', ``let me reconsider''), and considering an alternative (\eg ``or maybe'', ``let's assume''), which we call self-checks.
Since our models express the same move either in the native style of Qwen3.5-9B (``Let's check'') or in the first-person style of our SFT traces (``Let me check'', ``I need to''), each move covers both phrasings, and we normalize the count by the thinking length.
We further describe the allocation of the thinking budget as the ratio of the thinking length on hard questions to that on easy questions, where we rank the questions by the mean solve rate of five models\ifdefined\ppmode\space(the domain teachers trained directly from Qwen3.5-9B, their MOPD student, \oursmodel, MixSFT and the MixRL baseline, with easy questions at a solve rate of at least 0.9 and hard questions at most 0.25)\fi, the stability as the share of responses with at least five repeated lines, the decisiveness as the share of responses that hedge (\eg ``probably'', ``I'll go with''), and, on video, the grounding as the number of cited timestamps.
We compute all measures on the logged response of each question, which is the first of the eight samples of the avg@8 evaluation (Table~\ref{tab:behavior}).
\fi

\ifdefined\ppmode\else\fi

\paragraph{RL keeps the reasoning style of its start.}
In the RL runs of Table~\ref{tab:behavior}, the final model keeps the reasoning style of its start.
The models fall into two groups: Qwen3.5-9B and the models trained from it without SFT reason in the native style of Qwen3.5-9B, and the models trained on our SFT traces reason in the first-person style of these traces.
The phrasing stays the same in each group, since all models in the native style begin their thinking with a restatement of the task (``The user wants''), whereas no model in the first-person style does so, either before or after RL, and in the qualitative check, MixSFT and the MixRL baseline often begin with the same sentences and differ only at single steps (Table~\ref{tab:excerpts}).
RL instead changes the thinking length, the loops, and the way in which the model reaches its decision, which agrees with the finding that RL mainly sharpens behavior that the start model already shows~\cite{yue2025rlbeyondbase,gandhi2025cognitive}.

\begin{table*}[t]
\centering\footnotesize
\setlength{\tabcolsep}{4pt}
\renewcommand{\arraystretch}{1.18}
\caption{Excerpts from pairs of responses of two models to the same benchmark question. Each row compares the model before and after one stage, and (correct) and (wrong) give the benchmark verdict of the response.}
\label{tab:excerpts}
\begin{tabular}{@{}>{\raggedright\arraybackslash}p{0.15\textwidth}p{0.405\textwidth}p{0.405\textwidth}@{}}
\toprule
\textbf{Stage} & \textbf{Before} & \textbf{After} \\
\midrule
Domain RL\newline\textcolor{domImgr}{Image STEM} (DynaMath) &
\textbf{Qwen3.5-9B} (wrong): \textit{``Wait, let's look at the reflection again. \dots''} The same check returns until the response limit, and no answer is given. &
\textbf{Domain teacher} (correct): tries a rotation of the figure instead and checks it: \textit{``This matches the bottom text perfectly!''} \\
Domain RL\newline\textcolor{domVid}{Video} (Video-MME) &
\textbf{Qwen3.5-9B} (correct): walks through the clip: \textit{``Around 00:10, the first dive sequence starts. It shows a solo dive. Wait, let me look closer. At 00:10, we see one diver. Then at 00:13, \dots''} &
\textbf{Domain teacher} (correct): lists only the relevant moments: \textit{``Around 00:43, the two athletes are diving together again.''} \\
MOPD\newline\textcolor{domVid}{Video} (Video-MME) &
\textbf{Domain teacher} (correct): \textit{``The user wants to identify the location of the final scenes in the video. 1. Observe the video content from around 00:48 onwards.''} &
\textbf{MOPD student} (correct): \textit{``The user wants to identify the location of the final scenes in the video. 1. Look at the video from 00:48 onwards.''} \\
MixRL on top of MOPD\newline\textcolor{domVid}{Video} (Video-Holmes) &
\textbf{MOPD student} (wrong): builds a timestamped chain on a misperceived event: \textit{``At 01:38, a man with a gun is seen firing.''} &
\textbf{\oursmodel} (correct): states the event that the question asks about: \textit{``The video shows the woman splitting the wall with a sickle, creating an opening.''} \\
MixRL from MixSFT\newline\textcolor{domImgp}{Image perception} (CV-Bench) &
\textbf{MixSFT} (wrong): \textit{``I need to count the umbrellas in the image. Looking at the scene, I see a group of colorful umbrellas \dots''} &
\textbf{MixRL baseline} (correct): same opening, then one recount: \textit{``I need to count the umbrellas in the image. Looking at the scene, \dots\ Wait, let me recount carefully.''} \\
\bottomrule
\end{tabular}
\end{table*}

\paragraph{Domain RL.}
Qwen3.5-9B overthinks on all domains~\cite{chen2024overthinking}: its thinking on video is also long, part of its responses repeat the same check until the response limit, and on video, it visits many parts of the clip before it decides (Tables~\ref{tab:behavior} and~\ref{tab:excerpts}).
Domain RL keeps this style but sets a different thinking length for each domain, so the image STEM teacher still reasons at length, whereas the video teacher answers with a small fraction of the original length and cites fewer and more relevant moments of the clip.
At the same time, loops become much rarer, and in the qualitative check, the teacher tries another hypothesis and commits to an answer where Qwen3.5-9B checks the same detail again and again.

\paragraph{MOPD.}
The MOPD student keeps the behavior of each teacher not only on average (Table~\ref{tab:behavior}) but also per question: of all models, it is the closest to its teachers both in the questions that it solves and in its thinking length on each question (\mbox{Table}~\ref{tab:fidelity}).
The student of the teachers with the SFT warm-up agrees more with its own teachers than with the teachers trained directly from Qwen3.5-9B, so the student follows the specific teachers that it learns from rather than a common behavior of all RL models, and in the qualitative check, it often repeats the plan and even the wording of its teacher (Table~\ref{tab:excerpts}).
This fits on-policy distillation, which trains the student toward the teacher on the student's own responses~\cite{agarwal2024gkd,lu2025onpolicy}.

\begin{table}[t]
\centering\footnotesize
\setlength{\tabcolsep}{3.4pt}
\renewcommand{\arraystretch}{1.05}
\caption{Per-question agreement with the domain teachers trained directly from Qwen3.5-9B on the benchmarks of each domain: Spearman correlation of the number of correct samples out of eight (Solved) and of the thinking length of one response (Length), mean over the benchmarks. Dots: image STEM (\domdot{domImgr}), image perception (\domdot{domImgp}), video (\domdot{domVid}). $^\ast$Compared with its own teachers (with the SFT warm-up).}
\label{tab:fidelity}
\begin{tabular}{@{}lcccccc@{}}
\toprule
 & \multicolumn{3}{c}{Solved} & \multicolumn{3}{c}{Length} \\
\cmidrule(lr){2-4}\cmidrule(l){5-7}
\textbf{Model} & \domdot{domImgr} & \domdot{domImgp} & \domdot{domVid} & \domdot{domImgr} & \domdot{domImgp} & \domdot{domVid} \\
\midrule
Qwen3.5-9B & 0.84 & 0.80 & 0.77 & 0.81 & 0.69 & 0.30 \\
MixSFT & 0.69 & 0.71 & 0.76 & 0.58 & 0.44 & 0.22 \\
+ MixRL (MixRL baseline) & 0.80 & 0.77 & 0.81 & 0.71 & 0.58 & 0.32 \\
MOPD, with SFT warm-up & 0.81 & 0.77 & 0.82 & 0.73 & 0.57 & 0.31 \\
MOPD, without SFT warm-up & \textbf{0.90} & \textbf{0.87} & \textbf{0.91} & \textbf{0.86} & \textbf{0.77} & \textbf{0.60} \\
+ MixRL (\oursmodel) & 0.89 & 0.83 & 0.81 & 0.85 & 0.70 & 0.28 \\
\midrule
MOPD, with SFT warm-up$^\ast$ & 0.89 & 0.81 & 0.88 & 0.83 & 0.65 & 0.48 \\
\bottomrule
\end{tabular}
\end{table}

\paragraph{MixRL on top of MOPD.}
Besides the changes on video in Section~5\ofmain{}, MixRL on top of the MOPD student of the teachers trained directly from Qwen3.5-9B changes little on the two image domains, except that on image perception, the thinking length also depends more on the difficulty of the question.
On top of the MOPD student of the teachers with the SFT warm-up, MixRL also strengthens this dependence on video and adds self-checks, while the responses stay short (Table~\ref{tab:behavior}).

\paragraph{Qualitative check of reasoning traces.}
We compare pairs of responses of two models to the same benchmark question, sampled with more weight on questions where only one of the two responses is correct, and Table~\ref{tab:excerpts} shows one excerpt pair for each stage, with every excerpt verbatim.

\ifdefined\ppmode\else
\section{Gains on Each Benchmark}\label{sec:suppl_gains}

Figure~\ref{fig:gains} shows the gain of the two MOPD students, of MixSFT and of the MixRL baseline on each benchmark, which complements the results of Tables~1 and~2\ofmain{}.

\begin{figure*}[t]
  \centering
  \includegraphics[width=\textwidth]{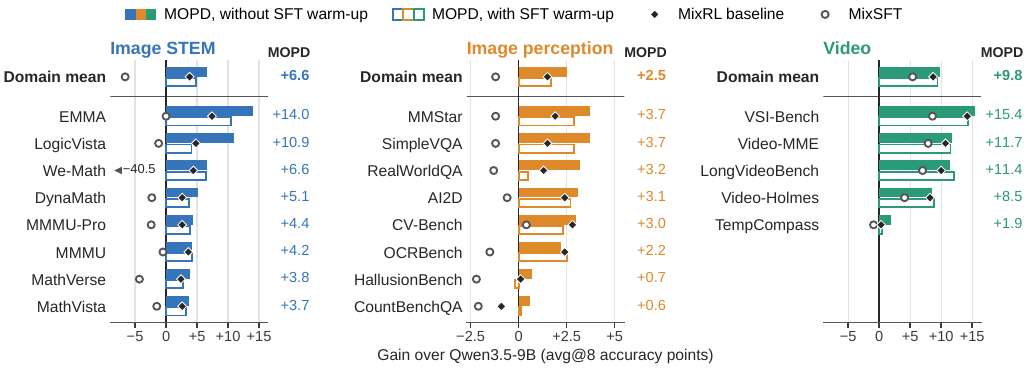}
  \caption{Gain over Qwen3.5-9B on each benchmark (avg@8 accuracy, points), with the domain mean first. Solid bars show MOPD of the teachers trained directly from Qwen3.5-9B (value on the right, which also sorts the benchmarks), and outlined bars show MOPD of the teachers with the SFT warm-up. An arrow with a value marks a gain outside the axis range. Both MOPD students improve on most benchmarks, and the MixRL baseline stays below them on most benchmarks.}
  \label{fig:gains}
\end{figure*}

\fi

\end{document}